\documentclass{article}

\PassOptionsToPackage{numbers,sort&compress}{natbib}
\usepackage[preprint]{neurips_2026}

\usepackage[utf8]{inputenc}
\usepackage[T1]{fontenc}
\usepackage{hyperref}
\usepackage{simpleicons}
\usepackage{url}
\usepackage{booktabs}
\usepackage{multirow}
\usepackage{longtable}
\usepackage{amsfonts}
\usepackage{nicefrac}
\usepackage{microtype}
\usepackage[table]{xcolor}
\usepackage{graphicx}
\usepackage{capt-of}
\usepackage{placeins}
\usepackage{amsmath}
\usepackage{amssymb}
\usepackage{tikz}
\usepackage{xspace}

\title{MoME: Mixture-of-Memory Embeddings for Context-Aware Sparse Lookup}

\author{
  \makebox[0.62\textwidth][c]{%
    Muchen Li\textsuperscript{1,2}
    \hfill
    Leonid Sigal\textsuperscript{1,2,3,4}
    \hfill
    Renjie Liao\textsuperscript{1,2,3}%
  } \\[1em]
  \textsuperscript{1}University of British Columbia \\[0.25em]
  \makebox[0.80\textwidth][c]{%
    \textsuperscript{2}Vector Institute for AI
    \hfill
    \textsuperscript{3}Canada CIFAR AI Chair
    \hfill
    \textsuperscript{4}NSERC CRC Chair%
  }
}

\newcommand{\ourfull}{Mixture of Memory Embeddings}
\newcommand{\ourabbr}{MoME}
\newcommand{\ours}{\ourabbr\xspace}

\newcommand{\eg}{\textit{e.g.}}

\begin{document}
\maketitle
\begingroup
\renewcommand{\thefootnote}{}
\makeatletter
\def\Hy@footnote@currentHref{Hfootnote.email}
\makeatother
\footnotetext{Contact: \texttt{muchenli@cs.ubc.ca}}
\endgroup

\begin{abstract}
Scaling large language models efficiently has motivated sparse capacity mechanisms such as Mixture-of-Experts and, more recently, conditional memory: token-indexed embedding tables that augment the backbone with cheap parametric lookups. Existing memory-embedding methods retrieve via a deterministic function of the surface form, which collapses different contextual senses of the same token (\eg, \emph{python} the language vs.\ the animal) into a single fixed entry. We introduce \ourfull{} (\ours{}), a context-aware memory mechanism that replaces each token's single memory row with a mixture of $M$ slots and uses a learned gate over the hidden state to choose which slots to read at each position. In controlled pretraining experiments across nanochat, Llama-3/MobileLLM, and Qwen3 backbones, \ours{} improves over Value Embedding, Bigram, and STEM baselines in iso-parameter and iso-training-FLOP settings, shows a more promising memory-size scaling trend at sub-billion scale, and remains efficient in training and inference. Qualitative routing analyses on polysemous tokens further suggest that the learned mixture exhibits a degree of semantic interpretability, dispatching the same surface token to distinct memory slots under different senses. The \href{https://github.com/jojo23333/Mixutre-Of-Memory-Embedding}{\mbox{\simpleicon{github}\,code}} and \href{https://huggingface.co/jojo23333/MOME-NanoChat-D24-100B}{\mbox{\raisebox{-0.15ex}{\includegraphics[height=1.15em]{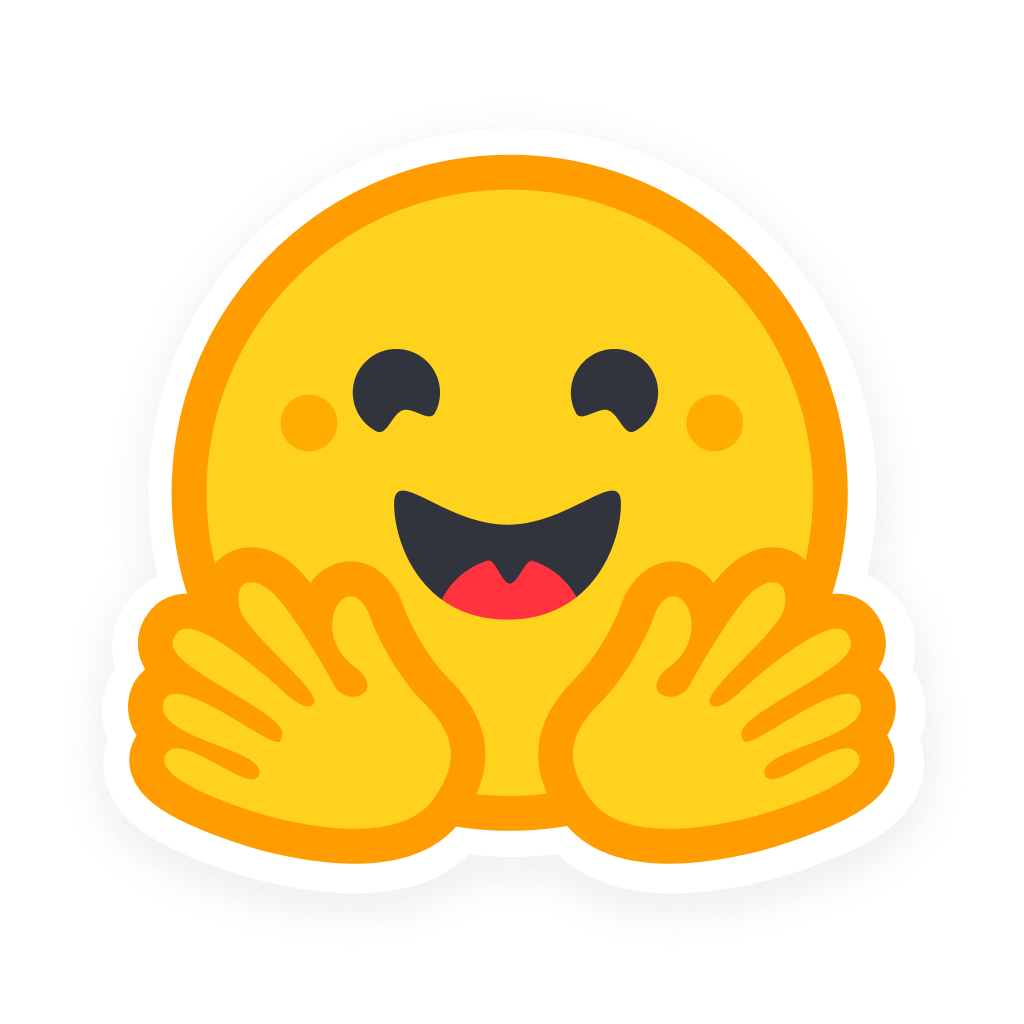}}\,pretrained models}} are open-sourced.
\end{abstract}

% Introduction
\section{Introduction}

Efficient scaling for large language models has been a central challenge for extending their capability boundary. Dense scaling improves performance~\citep{kaplan2020scaling,hoffmann2022training}, but it ties capacity growth to broadly active computation.
Mixture-of-Experts (MoE) addresses this by routing each token to only a subset of experts, enabling conditional computation, and is now standard in frontier models~\citep{shazeer2017outrageously,lepikhin2020gshard,fedus2022switch,du2022glam,jiang2024mixtral}.

More recently, a second axis of sparse scaling---conditional memory---has emerged as a complementary route to expanding model capacity~\citep{cheng2026conditional,sadhukhan2026stem}.
The motivation is that language modeling interleaves two sub-tasks: compositional reasoning, demanding dynamic computation, and retrieval of local, static patterns like named entities and formulaic phrases. Lacking a native lookup primitive, Transformers simulate retrieval through computation, spending early-layer capacity to reconstruct a static table.
Conditional memory instead handles such regularities through sparse lookups: each token retrieves only a few memory entries that inject useful priors into the backbone~\citep{weston2014memory,sukhbaatar2015end,lample2019large,berges2024memory,huang2024ultra}.
Rather than storing all useful associations only in dense transformer weights, memory-embedding methods learn auxiliary tables that can be looked up and injected into the backbone, including Per-Layer Embedding in Gemma~3~\citep{gemma3_2025}, value-stream memory variants~\citep{koszarsky2024valueembeddings,zhou2024value,modded_nanogpt_2024,karpathy2025nanochat}, STEM~\citep{sadhukhan2026stem}, Engram~\citep{cheng2026conditional}, and Bigram~\citep{classiclarry2026bigramhash}.
The appeal is efficiency: only a small fraction of the table is active for any token, the lookup is lightweight, and the table can therefore be scaled with modest additional compute.

These methods differ in where memory is injected, but they share a common restriction in how it is indexed. Existing memory-embedding methods usually retrieve memory through deterministic token or local $n$-gram indexing: Per-Layer Embedding uses token identity~\citep{gemma3_2025}; Value Embedding and related value-stream variants use token identity~\citep{koszarsky2024valueembeddings,zhou2024value,modded_nanogpt_2024,karpathy2025nanochat}; STEM uses token-indexed embedding modules~\citep{sadhukhan2026stem}; and Engram uses a fixed $n$-gram hash~\citep{cheng2026conditional}. Broader work on embedding, vocabulary, and $n$-gram scaling further motivates this direction~\citep{yu2025scaling,huang2025over,tao2024scaling,roy2022ngrammer,liu2024infinigram}. This makes retrieval simple, but it also makes the retrieved memory largely context-blind. The same token can call for different associations in different contexts: \emph{python} may refer to a programming language or an animal, and \emph{spring} may refer to a season, a mechanical coil, or a verb. A single deterministic memory vector for such tokens forces these contextual modes to share one vector.

To address this shortcoming, we introduce \ourfull{} (\ours{}), a {\em context-aware} conditional memory mechanism that is more expressive but equally efficient, retaining the access pattern of token-indexed lookups.
%   
% In this paper, we introduce context-aware memory retrieval through \ourfull{} (\ours), a conditional memory mechanism that breaks the context-blind nature of token-indexed lookups while preserving their efficient access pattern.
% 
Instead of assigning each token row a single fixed memory vector, \ours stores a mixture of memory slots and uses the current hidden context to choose which components of that mixture to read. This gives memory retrieval a simple form of contextual adaptivity while keeping the mechanism close to the efficient lookup structure used by prior memory embeddings.

We evaluate \ours in controlled pretraining experiments across three architecture families: nanochat-style~\citep{karpathy2025nanochat}, Llama~3/MobileLLM-style~\citep{grattafiori2024llama3herdmodels,liu2024mobilellmoptimizingsubbillionparameter}, and Qwen3-style backbones~\citep{yang2025qwen3technicalreport}.
Across these settings, \ours improves over comparable memory-augmented transformer baselines in most matched comparisons, including Base, Value Embedding~\citep{koszarsky2024valueembeddings,zhou2024value}, STEM~\citep{sadhukhan2026stem}, and Bigram~\citep{classiclarry2026bigramhash} baselines. 
We also study memory-size scaling on the nanochat-style backbone in a low-compute regime, where \ours shows a more favorable observed scaling trend than Bigram~\citep{classiclarry2026bigramhash} over the tested memory range. A complementary ablation further indicates that \ours{} is compatible with Bigram under compound scaling.
In addition, qualitative and quantitative routing analyses on a curated set of polysemous tokens (e.g., \emph{apple}, \emph{bank}, \emph{python}) show that \ours dispatches the same token to distinct memory slots under different semantic contexts on both nanochat- and Qwen3-style backbones, indicating that the learned mixture captures contextual variation rather than committing to one route per token.

Our contributions are:
\begin{itemize}
  \item We design \ours, a mixture-of-memory module that brings context awareness into the memory table lookup of token-indexed memory embeddings.
  \item We show that \ours achieves competitive performance against prior state-of-the-art memory-embedding methods and transfers well across diverse model architectures.
  \item We further show that \ours exhibits a more favorable memory-size scaling trend than existing baselines, suggesting better returns as memory capacity grows; additionally, \ours can be applied together with Bigram to achieve better performance.
  \item Routing analyses show that the proposed module learns to dynamically index different memory slots in context, with the selected slots reflecting the semantics of the input.
\end{itemize}

% Related Work
\section{Related Work}\label{sec:related-work}

\paragraph{Learning memories of Large Language Models.}\label{sec:rw-memory}
Memory-augmented LMs span non-parametric retrieval~\citep{guu2020realm, borgeaud2022retro} and end-to-end memory networks~\citep{weston2014memory, sukhbaatar2015end}. Both are orthogonal to \ours{}, which keeps memory parametric and token-indexed. Closer to our setting, FFN layers behave as key-value memories~\citep{geva2021transformer, dai2022knowledge, meng2022locating}, motivating explicit memory layers with sub-linear lookup~\citep{lample2019large, berges2024memory, huang2024ultra}. \ours{} shares this parametric premise but injects into the per-head value stream and indexes by token identity, trading content-addressable lookup for a lighter retrieval path.

\paragraph{Memory-Embedding Augmented LLMs.}\label{sec:rw-memory-emb}
A recent convergent line attaches a learnable embedding table to the transformer and looks it up by a deterministic function of the input. Per-Layer Embedding in Gemma~3n~\citep{gemma3_2025} uses per-token-id rows at each layer's input; Value Embedding~\citep{modded_nanogpt_2024, karpathy2025nanochat, koszarsky2024valueembeddings, zhou2024value} attaches the table to the value stream; STEM~\citep{sadhukhan2026stem} fuses a token-indexed lookup into the SwiGLU hidden state; Engram~\citep{cheng2026conditional} uses deterministic $n$-gram lookup followed by hidden-state-conditioned fusion; our Bigram baseline~\citep{classiclarry2026bigramhash} is a separate implementation using only 2-gram lookup; MoWE~\citep{santos2024mowe} routes tokens deterministically into many small word-experts. Across this family, retrieval is a fixed function of the surface form, not of the hidden state -- the gap \ours{} closes via a context-aware mixture-of-slots while keeping the cheap lookup.
A parallel sub-line scales the embedding table itself. SCONE~\citep{yu2025scaling} learns frequent-n-gram embeddings via an auxiliary contextualizer and offloads them at inference; Over-Tokenized Transformer~\citep{huang2025over} feeds many overlapping n-gram embeddings, and Byte Latent Transformer~\citep{pagnoni2025blt} does so at the byte level. These extend earlier modular-hash precursors~\citep{roy2022ngrammer, huang2021lookup, liu2024infinigram} and are motivated by vocabulary scaling laws~\citep{tao2024scaling}. Routing here also remains a fixed function of the surface form.
\ours{}'s value-stream injection relates to cross-layer value designs: ResFormer~\citep{zhou2024value} adds a residual from the first layer's value vectors; NeuTRENO~\citep{nguyen2023neutreno} regularizes value vectors across layers; DenseFormer~\citep{pagliardini2024denseformer} depth-weight-averages hidden states; Cross-Layer Attention variants~\citep{brandon2024cla, mu2024crosslayer} share K/V across layers. \ours{} chooses this site so the memory branch runs in parallel with the value projection, joining only at a gated residual addition (Section~\ref{sec:injection}).

\paragraph{Mixture of Experts.}\label{sec:rw-moe}
\ours{}'s slot gate borrows from Mixture-of-Experts: the original sparsely-gated MoE~\citep{shazeer2017outrageously} and its descendants~\citep{lepikhin2020gshard, fedus2022switch, du2022glam, jiang2024mixtral} decoupled capacity from per-token FLOPs. Closer to our setting are fine-grained designs that split each FFN into many small specialists with optional shared experts~\citep{dai2024deepseekmoe, liu2024deepseekv3, he2024mixture}. \ours{} applies the same idea on the \emph{memory-table} side: each token-indexed row holds $M$ slots, and the gate selects a subset at the current position.

% MoE routing also offers useful design references. Hash Layers use fixed routing~\citep{roller2021hash}, Expert-Choice reverses the assignment direction~\citep{zhou2022expertchoice}, and auxiliary-loss-free bias updates provide a lightweight load-balancing mechanism~\citep{wang2024lossfree}; \ourfull{} adopts this bias-update idea for memory-slot routing.

\section{\ourfull{}}\label{sec:method}

\subsection{\texorpdfstring{Preliminaries -- Token-Indexed Table as Learnable Memory}{Preliminaries -- Token-Indexed Table as Learnable Memory}}\label{sec:prior-memory}

% Comparison of how prior works insert memory embeddings.
\begin{table*}[!t]
  \caption{Comparison of how prior works retrieve and inject memory embeddings. Context aggregation describes memory addressing; canonical Engram additionally conditions fusion on the hidden state.}
  \label{tab:related-memory}
  \centering
  \footnotesize
  \setlength{\tabcolsep}{4pt}
  \renewcommand{\arraystretch}{1.2}
  \begin{tabular}{@{}lclll@{}}
    \toprule
    Method & Memory index & Index function & Injection position & Context aggregation \\
    \midrule
    Per-Layer Embedding~\citep{gemma3_2025} & $V \times d_{\mathrm{model}}$        & $x_t$                                      & Input embedding & None \\
    Value Embedding~\citep{zhou2024value}   & $V \times H \times d_{\mathrm{value}}$              & $x_t$                                      & Attention value & None \\
    STEM~\citep{sadhukhan2026stem}          & $V \times d_{\mathrm{ffn}}$          & $x_t$                                      & SwiGLU hidden   & None \\
    Engram (canonical)~\citep{cheng2026conditional}     & $N_{\text{hash}} \times d_{\mathrm{model}}$ & $h(x_t, \dots , x_{t-n})$          & Block input & $n$-gram \\
    \midrule
    \textbf{Ours}                           & $N \times M \times d_{\mathrm{value}}$              & $\bigl[f(x_t),\; \mathcal{A}_{t,i}\bigr]$ & Attention value & Hidden embedding \\
    \bottomrule
  \end{tabular}
\end{table*}

Prior works on memory-augmented transformers~\citep{gemma3_2025, zhou2024value, sadhukhan2026stem, cheng2026conditional} (Table~\ref{tab:related-memory}) can be written as learning an embedding table $\mathbf{E}\in\mathbb{R}^{N \times D_{\mathrm{inj}}}$, where $N$ is the number of addressable memory entries and $D_{\mathrm{inj}}$ is the dimensionality required by the injection site.
For each input token $x_t$, the memory table is indexed deterministically, either directly by token id ($N=V$) or by a hash function over an $n$-gram window ($N=N_{\text{hash}}$).
These works further differ in where this retrieved embedding is injected into the backbone. Per-Layer Embedding~\cite{gemma3_2025} injects at the input embedding, Engram~\cite{cheng2026conditional} fuses retrieved memory into block-input hidden states through a context-dependent gate, value embedding~\cite{zhou2024value,modded_nanogpt_2024,karpathy2025nanochat} injects into the attention value stream, and STEM fuses the retrieved embedding into the SwiGLU hidden state~\citep{shazeer2020glu}. Table~\ref{tab:related-memory} organizes the representative methods along these axes; across the space the lookup remains a deterministic function of $x_t$ alone (or of a fixed $n$-gram), so memory addressing does not depend on the model's hidden state at position $t$, even when fusion is context-dependent.

We argue that token-deterministic indexing has one underlying mismatch: capacity is allocated to tokens rather than to semantic content. This manifests as two compounding drawbacks:

\begin{list}{}{%
  \setlength{\topsep}{1pt}%
  \setlength{\partopsep}{0pt}%
  \setlength{\parsep}{0pt}%
  \setlength{\itemsep}{2pt}%
  \setlength{\leftmargin}{1.4em}%
  \setlength{\labelwidth}{1.0em}%
  \setlength{\labelsep}{0.3em}%
  \setlength{\rightmargin}{0pt}%
}
% \item[\textbf{(i)}] \textbf{Inefficient Learning.}
% To begin with. Token frequency under standard subword tokenizers follows a Zipfian distribution where a small number of high-frequency tokens account for most occurrences, while most token types appear rarely. therefore the learning of memory tables are naturally imbalanced: a small head of memory slots accounts for the bulk of pretraining tokens while the long tail is visited only a handful of times.
% Moreover,
% 

\item[\textbf{(i)}] \textbf{Misallocated capacity.} One slot per token treats all tokens as if they carried equal semantic load, but they do not. Some tokens are near-redundant: variants like \emph{cats}/\emph{cat} largely overlap in meaning, yet each occupies its own row and the table duplicates the same content; other tokens are the opposite---a single row is asked to hold several distinct meanings, e.g., \emph{bank} (financial institution vs.\ river edge), with no room to keep them apart.

\item[\textbf{(ii)}] \textbf{Context-blind retrieval.} Deterministic lookup makes memory retrieval blind to the full context.\footnote{Canonical Engram uses 2- and 3-gram lookup; our Bigram baseline uses only 2-grams.} Token semantics vary substantially across contexts---\emph{python} as a programming language versus an animal---so memory for a token should be multi-modal across contexts, yet a deterministic table dedicates a single memory slot per token.
\end{list}

\subsection{\texorpdfstring{Learning \ourfull{}}{Learning \ourfull}}\label{sec:memory-construction}

\paragraph{Designing \ourfull{} Tables.}
Motivated by the limitations of current memory designs, we introduce \ourfull{}, which extends the prior token-indexed memory tables to resolve the limitation mentioned in Section~\ref{sec:prior-memory}.

Specifically, we introduce two designs: (i) \textbf{A mixture of $M$ memory slots with a learned context-aware gate.} Motivated by Mixture-of-Experts models~\citep{shazeer2017outrageously}, we extend each token-indexed row to $M$ memory slots with a learned context-aware gate to selectively activate the memory embedding. (ii) \textbf{Grouped token indexes $f$.} Aiming to reduce the redundancy in the memory table, we introduce an optional deterministic function to group tokens with similar semantics together. We find that this improves training efficiency.

Concretely, the memory is a learnable parameter tensor
\[
\mathbf{E}^{\mathrm{mem}} \in \mathbb{R}^{N \times M \times d_{\mathrm{value}}},
\]
where $N$ is the number of rows selected by the first-stage indexer $f$ ($N=V$ for identity indexing and $N\le V$ after grouping), $M$ is the number of slots per row, and $d_{\mathrm{value}}$ is the per-head value dimension at the memory injection point. We write the stored memory vector at row $n$ and slot $a$ as
\[
\mathbf{m}_{n,a}
= \mathbf{E}^{\mathrm{mem}}[n,a,:]
\in\mathbb{R}^{d_{\mathrm{value}}}.
\]
% The choice of $d_{\mathrm{value}}$ reflects that \ourfull{} injects memory separately into each value head rather than constructing one full $H d_{\mathrm{value}}$-dimensional vector per slot. Each $\mathbf{m}_{n,a}$ has exactly the shape needed to be added to a single value-head vector $\mathbf{v}_{t,i}$.
%

\begin{figure*}[!t]
  \centering
  \includegraphics[width=\textwidth]{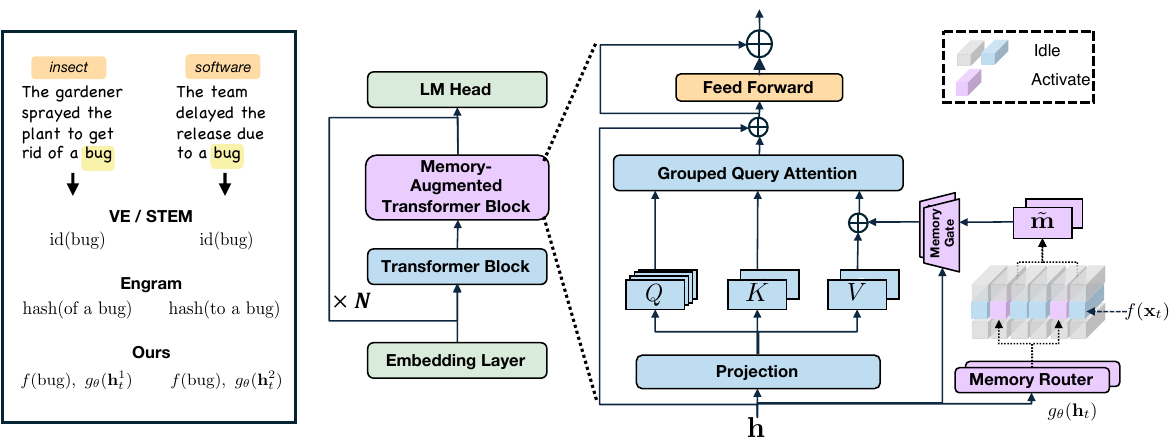}
  \caption{\textbf{Left:} the same token can carry different senses, and prior token-indexed methods (VE/STEM, canonical Engram) commit each row to a single fixed memory while \ours{} dispatches to different slots depending on the hidden state (see Figure~\ref{fig:main-semantic-routes} and Section~\ref{sec:appendix-semantic-routes} for the learned routing). \textbf{Right:} \ours{} alternates regular transformer blocks with memory-augmented blocks; in each memory-augmented block, a context-aware router retrieves a memory vector and injects it into the attention value stream.}
  \label{fig:pipeline}
\end{figure*}

Figure~\ref{fig:pipeline} shows the overall \ourfull{} architecture: regular transformer blocks alternate with memory-augmented transformer blocks. Each memory-augmented block keeps the standard transformer computation and attaches a side memory module: the module reads the block input, retrieves a context-dependent memory vector, and injects it back into the attention value stream for each value head. 
Here, $x_t$ denotes the input token at position $t$, $\mathbf{h}_t$ denotes the hidden state entering the memory-augmented block, and $\mathcal{A}_{t,i}$ denotes the active slot set selected for value head $i$.
We also show the two memory indexers explicitly: $f(x_t)\to n_t$ selects the memory row, and $g_\theta(\mathbf{h}_t)\to(\mathcal{A}_{t,i},\alpha)$ selects and weights slots within that row before value injection.

\subsection{\texorpdfstring{Context-Aware Routing and Aggregation}{Context-Aware Routing and Aggregation}}\label{sec:gate}
\paragraph{Context-Aware Routing.}
\ourfull{} can be viewed as composing two memory indexers. The token indexer $f(x_t)$ selects the row $n_t$, while the context indexer $g_\theta(\mathbf{h}_t)$ selects and weights slots within that row. Given $n_t=f(x_t)$, the second-stage gate decides which of the $M$ slots in that row are selected at the current position. This is where context enters the lookup: rather than committing the row to a single fixed vector, a learned gate over the hidden state selects the memory slots dynamically.
For each value head $i\in[H]$, the gate is implemented by first mapping the hidden state to slot logits
\[
\boldsymbol{\ell}_{t,i}
= \mathbf{W}^{(i)}_g\mathbf{h}_t+\mathbf{b}^{(i)}_g
\in\mathbb{R}^{M}.
\]
Here $\boldsymbol{\ell}_{t,i}$ are the routing logits used for memory-slot selection.
% The memory tensor $\mathbf{E}^{\mathrm{mem}}$ is shared across heads; only the gate output is head-specific, letting each head select its own subset of slots within the shared row. 
Conditioning the slot gate on $\mathbf{h}_t$ is what makes routing context-aware, since $\mathbf{h}_t$ has already integrated context through the upstream transformer layers.
\[
\mathcal{A}_{t,i}
= \mathrm{TopK}(\boldsymbol{\ell}_{t,i}, K),
\qquad
\mathcal{A}_{t,i}\subseteq[M],
\qquad
\lvert\mathcal{A}_{t,i}\rvert=K.
\]

\paragraph{Memory Aggregation.}
To aggregate memory, we use the sigmoid-norm gate for $K>1$, converting slot logits into scores and normalizing only over the selected slots
\[
\mathbf{s}_{t,i}=\sigma(\boldsymbol{\ell}_{t,i}),\qquad
\alpha^{(i)}_{t,a}
= \frac{s_{t,i,a}}{\sum_{b \in \mathcal{A}_{t,i}} s_{t,i,b}},
\quad a\in\mathcal{A}_{t,i}.
\]
For $K=1$, we use the softmax variant inline, $\alpha^{(i)}_{t,a}=[\mathrm{softmax}(\boldsymbol{\ell}_{t,i})]_a$.
% with $\alpha^{(i)}_{t,a}=0$ for $a\notin\mathcal{A}_{t,i}$. Equivalently, for each head we write the full slot gate as
% \[
% g_\theta^{(i)}(\mathbf{h}_t)
% =
% \left(
% \mathcal{A}_{t,i},
% \{\alpha^{(i)}_{t,a}\}_{a\in\mathcal{A}_{t,i}}
% \right).
% \]
% The gate $g_\theta$ therefore produces both the activated slot set and the mixture weights used to aggregate memory from the row selected by $f(x_t)$.

Given $n_t = f(x_t)$ and the mixture weights $\{\alpha^{(i)}_{t,a}\}_{a \in \mathcal{A}_{t,i}}$, the aggregated memory for head $i$ is constructed as the weighted sum of activated stored memory vectors:
\[
\tilde{\mathbf{m}}_{t,i}
= \sum_{a \in \mathcal{A}_{t,i}}
\alpha^{(i)}_{t,a}
\mathbf{m}_{n_t,a}
\in \mathbb{R}^{d_{\mathrm{value}}}.
\]

\subsection{Token-Index Grouping Function}\label{sec:token-index-grouping}
Optionally, training efficiency can be further improved by optimizing the token indexer itself: tokens with similar semantics are grouped into shared rows, reducing redundant token-indexed capacity while preserving context-aware slots within each row. We design the token-based indexer as a function
\[
f : \mathcal{V} \to [N], \qquad N \le V,
\]
that maps tokens with overlapping semantics into the same row. We tried several similarity heuristics and found that embedding-based matching works best: we instantiate $f$ from a lightly pretrained token-embedding matrix $\mathbf{U}^{\mathrm{ref}} \in \mathbb{R}^{V \times d_{\mathrm{value}}}$ (obtained from a baseline run with $f = \mathrm{id}$) by offline kNN matching in pretrained embedding space,
\[
f(x) = \mathrm{kNN}\big(\mathbf{U}^{\mathrm{ref}}_x;\, \mathbf{U}^{\mathrm{ref}},\, c_{\mathrm{grp}}\big),
\]
yielding a fixed slot map of grouping factor $c_{\mathrm{grp}}$ (\eg, $c_{\mathrm{grp}}{=}2,4$ in our main configurations) that is frozen for the duration of training; here $\mathrm{kNN}(\cdot)$ denotes the offline row-map construction, not a runtime retrieval. Grouping shrinks the token-indexed row dimension by $c_{\mathrm{grp}}$, and the saved capacity is offloaded to context-aware memory by increasing the number of slots $M$ per row by $c_{\mathrm{grp}}$.

\subsection{Memory Injection}\label{sec:injection}
\paragraph{Gated Injection.}
The aggregated memory $\tilde{\mathbf{m}}_{t,i}$ is fused into the attention value stream through a separate per-head value-residual gate
\[
\boldsymbol{\gamma}_{\theta}(\mathbf{h}_t)
= 2\sigma(\mathbf{W}_\gamma\mathbf{h}_t+\mathbf{b}_\gamma)
\in \mathbb{R}^{H},
\]
\[
\tilde{\mathbf{v}}_{t,i}
= \mathbf{v}_{t,i}
+ \gamma_{\theta,t,i}\tilde{\mathbf{m}}_{t,i},
\qquad
\gamma_{\theta,t,i}=[\boldsymbol{\gamma}_{\theta}(\mathbf{h}_t)]_i.
\]
Thus $g_\theta$ controls which memory slots are aggregated, while $\boldsymbol{\gamma}_{\theta}$ controls how strongly the aggregated memory is added to the value stream. The factor of $2$ makes $\gamma_{\theta,t,i}=1$ when the gate logit is initialized at zero, giving a neutral initial value-residual scale.

\paragraph{Latency-Friendly Design.}
The memory branch introduces extra per-head computation: the router projects hidden states to slot scores, performs top-$K$ selection, and aggregates the selected memory vectors. Although this branch is lightweight in FLOPs, it can still add inference latency because the additional kernels sit on the critical path. A benefit of conditioning on $\mathbf{h}_t$ and injecting into the value stream is that much of the memory branch can be scheduled in parallel with the standard value projection that produces $\mathbf{v}_{t,i}$. After both branches finish, the only required join is the gated residual addition that produces $\tilde{\mathbf{v}}_{t,i}$. In the \texttt{qwen3\_4b} two-memory-layer benchmark, this design adds $0.509$ ms, while the hidden-state-injection variant of \ours{} adds $0.996$ ms, nearly doubling the latency overhead. The headline numbers across backbones are summarized in Table~\ref{tab:inference-latency}; the full per-backbone latency sweep is in Appendix Table~\ref{tab:appendix-inference-latency}.

\paragraph{Multi-Head Memory.}
As a parameter-efficient design choice, all value heads in a memory-augmented layer share a single memory table $\mathbf{E}^{\mathrm{mem}}$, capping the per-layer memory parameter count at $N M d_{\mathrm{value}}$ rather than $N M H d_{\mathrm{value}}$. To retain head-specific aggregation flexibility under this shared table, the slot router/gate output is per-head: each value head independently selects its activated slot subset and mixture weights over the shared row, so different heads can disagree on which slots to read while drawing from the same stored bank.
% The prior single-row abstraction is recovered as the degenerate case $M=1$ with identity indexing $N=V$; our main configurations use $N < V$ and $M > 1$, distributing capacity across both axes.

% Results
\section{Experiments and results}
% we should use 4 digit precision here
\newcommand{\metricstd}[2]{$#1_{\tiny{\pm #2}}$}
\newcommand{\bestmetricstd}[2]{\textbf{\boldmath\metricstd{#1}{#2}}}
\newcommand{\secondmetricstd}[2]{\underline{\metricstd{#1}{#2}}}

\begin{table*}[!t]
  \caption{Pretraining results on the nanochat-style backbone. \textit{Memory} shape uses the notation in Table~\ref{tab:notation}; \textit{Param (M/T)} is memory-only / total parameters.}
  \label{tab:pretrain-12layer}
  \centering
  \scriptsize
  \setlength{\tabcolsep}{2.3pt}
  \renewcommand{\arraystretch}{1.15}
  \resizebox{\textwidth}{!}{%
  \begin{tabular}{@{}l c ccc cc@{}}
    \toprule
    Run & Memory & Train bpb $\downarrow$ & Val bpb $\downarrow$ & CORE $\uparrow$ & Param (M/T) & Train Throughput \\
    \midrule
    Base                       & --                                                                       & \metricstd{0.8783}{0.0009} & \metricstd{0.8785}{0.0010} & \metricstd{0.1447}{0.0064} & 0M / 135M    & 1.15M tok/s \\
    Bigram                     & $N_{\text{hash}} \times d_{\mathrm{model}}$                              & \metricstd{0.8632}{0.0009} & \metricstd{0.8636}{0.0009} & \metricstd{0.1533}{0.0116} & 151M / 286M  & 1.09M tok/s \\
    VEmbedding            & $V \times 6 \times d_{\mathrm{value}}$                                    & \metricstd{0.8631}{0.0006} & \metricstd{0.8633}{0.0006} & \metricstd{0.1522}{0.0045} & 151M / 286M  & 1.10M tok/s \\
    % Ours-A1/6-Softmax          & $V \times M \times d_{\mathrm{value}}$                & \metricstd{0.8626}{0.0003} & \metricstd{0.8628}{0.0003} & \metricstd{0.1563}{0.0025} & 151M / 286M  & -- \\
    % Ours-A2/12-factor2-Softmax & --                                                                     & \metricstd{0.8614}{0.0008} & \metricstd{0.8615}{0.0008} & \metricstd{0.1526}{0.0030} & 151M / 286M  & -- \\
    \midrule
    \multicolumn{7}{@{}l}{\textit{151M memory}} \\
    \rowcolor{blue!8}
    \ours ($c_{\mathrm{grp}}{=}1$)  & $V \times 6 \times d_{\mathrm{value}}$              & \metricstd{0.8618}{0.0002} & \metricstd{0.8621}{0.0003} & \metricstd{0.1571}{0.0030} & 151M / 286M & 1.07M tok/s \\
    \rowcolor{blue!8}
    \ours ($c_{\mathrm{grp}}{=}2$) & $\frac{V}{2} \times 12 \times d_{\mathrm{value}}$   & \secondmetricstd{0.8609}{0.0003} & \secondmetricstd{0.8611}{0.0003} & \bestmetricstd{0.1583}{0.0020} & 151M / 287M & 1.07M tok/s \\
    \rowcolor{blue!8}
    \ours ($c_{\mathrm{grp}}{=}4$) & $\frac{V}{4} \times 24 \times d_{\mathrm{value}}$   & \bestmetricstd{0.8608}{0.0004} & \bestmetricstd{0.8610}{0.0004} & \metricstd{0.1554}{0.0029} & 151M / 287M & 1.07M tok/s \\
    \midrule
    \multicolumn{7}{@{}l}{\textit{302M memory}} \\
    \rowcolor{blue!8}
    \ours ($c_{\mathrm{grp}}{=}1$) & $V \times 12 \times d_{\mathrm{value}}$             & \metricstd{0.8567}{0.0002} & \metricstd{0.8569}{0.0002} & \secondmetricstd{0.1635}{0.0067} & 302M / 438M & 1.03M tok/s \\
    \rowcolor{blue!8}
    \ours ($c_{\mathrm{grp}}{=}2$) & $\frac{V}{2} \times 24 \times d_{\mathrm{value}}$   & \bestmetricstd{0.8558}{0.0003} & \bestmetricstd{0.8561}{0.0003} & \bestmetricstd{0.1664}{0.0059} & 302M / 438M & 1.03M tok/s \\
    \rowcolor{blue!8}
    \ours ($c_{\mathrm{grp}}{=}4$) & $\frac{V}{4} \times 48 \times d_{\mathrm{value}}$   & \secondmetricstd{0.8563}{0.0005} & \secondmetricstd{0.8565}{0.0006} & \metricstd{0.1515}{0.0110} & 302M / 439M & 1.02M tok/s \\
    \bottomrule
  \end{tabular}
  }
\end{table*}

% MobileNet pretraining summary moved to tables/mobilenet_result.tex.
% Qwen3 pretraining summary moved to tables/qwen3_result.tex.

\subsection{\texorpdfstring{Evaluation Setting}{Evaluation Setting}}\label{sec:eval-setting}

\textbf{Evaluation setting.} We follow the nanochat evaluation scripts~\citep{karpathy2025nanochat}\footnote{Our nanochat reference point is commit \texttt{348fbb3}.} and report final train bpb, validation bpb, raw benchmark accuracies, and the CORE metric~\citep{li2024datacomp}. 
We compute $\mathrm{bpb}=\frac{\sum_i-\log p_\theta(y_i\mid x_{<i})}{\log(2)\sum_i\mathrm{bytes}(y_i)}$ over counted target tokens, and $\mathrm{CORE}=\frac{1}{J}\sum_{j=1}^{J}\frac{a_j-r_j}{1-r_j}$ over task accuracies $a_j$ with task-specific random baselines $r_j$.
Lower bpb is better and higher CORE is better; 
Appendix~\ref{sec:appendix-eval-protocol} gives the detailed evaluation suite, and Appendix~\ref{sec:appendix-training-setup} gives the shared training setup.
For compactness, the main tables report representative raw accuracies and the CORE aggregate; see Appendix~\ref{sec:appendix-eval-protocol} for the full evaluation suite and per-task definitions, with the complete 22-task Llama/MobileLLM-family and Qwen3-style breakdowns in Appendix Table~\ref{tab:appendix-token-matched-full}.

\subsection{\texorpdfstring{Experiment Details}{Experiment Details}}\label{sec:experiment-details}
\textbf{Training settings.} For the experiments in Tables~\ref{tab:pretrain-12layer} and~\ref{tab:mobilenet125-summary}, we train on FineWeb-Edu data~\citep{lozhkov2024finewebedu} with 524{,}288-token batches. The compound-scaling runs in Table~\ref{tab:engram-mom-compound-scaling} use the same data and batch size. Our 100B-token runs in Table~\ref{tab:d24-100b-result} and the d24 control experiments in the appendix use ClimbMix~\citep{diao2025nemotronclimb} with 1{,}048{,}576-token batches. All use 2048-token sequences and the shared 32k BPE tokenizer. Training uses Muon for matrix-shaped transformer parameters and AdamW parameter groups for embeddings, unembeddings, scalars, value-memory tables, and other non-matrix parameters. Appendix~\ref{sec:appendix-training-setup} summarizes the concrete training setup used by the reported families.

\textbf{Base architectures.} We evaluate three architecture families: a nanochat-style GPT family~\citep{karpathy2025nanochat}, a Llama/MobileLLM-family setting~\citep{grattafiori2024llama3herdmodels,liu2024mobilellmoptimizingsubbillionparameter} used for the STEM comparison~\citep{sadhukhan2026stem}, and a Qwen3-style 0.6B family~\citep{yang2025qwen3technicalreport}.
The nanochat family is a compact decoder-only Transformer setting used for controlled iso-FLOP and iso-parameter comparisons, while the Llama/MobileLLM and Qwen3 families test whether the same memory mechanism transfers to stronger sub-billion-parameter backbone recipes.
All three families use the same 32k-token tokenizer trained on FineWeb-Edu with 20B tokens using byte-level BPE~\citep{radford2019language}.

\textbf{Baseline implementations.} We compare with Bigram, the best-performing Engram~\citep{cheng2026conditional} variant tested at our scale, using the modded-nanogpt 2-gram implementation~\citep{classiclarry2026bigramhash}. In our controlled comparison, it achieves lower validation bpb and higher CORE-22 than our canonical Engram adaptation (Appendix~\ref{sec:appendix-baseline-implementation}).
For STEM, we follow the original STEM paper and its released code~\citep{sadhukhan2026stem}; we use the same Llama/MobileLLM-family backbone for fair comparison and adopt the 1/2 STEM setting reported in their paper to align more closely with our setup.
Value embedding follows the value-embedding line from modded-nanogpt/nanochat practice~\citep{koszarsky2024valueembeddings,karpathy2025nanochat} and the value residual learning formulation~\citep{zhou2024value}. Appendix~\ref{sec:appendix-baseline-implementation} gives baseline implementation notes.

\textbf{Implementation details for our method.} For our method, we default to a single memory block per layer and alternate memory-augmented blocks and transformer blocks. Each memory block contains a memory table of size $N\times M\times d_{\mathrm{value}}$, where by default we set $M = H$ for an iso-parameter setting against value embedding.
We denote configurations as \ours{}-A$K$/$M$, where the prefix \emph{A} stands for \emph{Activated}: $K$ activated slots are selected out of $M$ slots per row (\eg, \ours{}-A2/5 activates $K{=}2$ slots out of $M{=}5$). Unless otherwise specified, the default activated-slot count is $K{=}2$. By default we do \emph{not} use the kNN row-grouping function, fixing $f(x_t){=}\mathrm{id}(x_t)$ and $c_{\mathrm{grp}}{=}1$ to isolate the effect of the memory mixture itself; the only place we sweep $c_{\mathrm{grp}}$ is the nanochat ablation in Section~\ref{sec:results-nanochat}, where the reference embedding table for kNN grouping is taken from a lightly trained nanochat-d12 base run ($\sim$1B tokens, $\sim$1e18 FLOPs). Appendix~\ref{sec:appendix-mome-family-configs} lists the run-specific \ours{} configurations across backbone families.
\subsection{\texorpdfstring{Pretraining Results}{Pretraining Results}}\label{sec:results-nanochat}
% Compact iso-training-token summary (Base vs. STEM/VE vs. Ours).
\begin{table*}[!t]
  \caption{Iso-training-token results for the Llama/MobileLLM and Qwen3 backbone families. Rows are grouped by backbone scale. \textit{Mem} is the memory-table size, per-benchmark columns report raw accuracy (\%), \textit{BB-avg} is the unweighted mean of the BigBench tasks in our suite, and \textit{Wall} is training time relative to the dense baseline within the same block. Full 22-task results appear in Table~\ref{tab:appendix-token-matched-full}.}
  \label{tab:mobilenet125-summary}
  \centering
  \scriptsize
  \setlength{\tabcolsep}{3pt}
  \renewcommand{\arraystretch}{1.2}
  \resizebox{\textwidth}{!}{%
  \begin{tabular}{@{}l c | cc | cccccccc | c@{}}
    \toprule
    Method & Mem & Val bpb $\downarrow$ & CORE $\uparrow$ & ARC-E & ARC-C & BoolQ & PIQA & HSwag & OBQA & WinoG & BB-avg & Wall \\
    \midrule
    \multicolumn{13}{@{}l}{\textit{Llama/MobileLLM 125M}} \\
    \midrule
    Base        & 0M   & 0.8852 & 0.1382 & 52.82 & 27.56 & \underline{55.14} & 63.82 & 33.27 & 33.40 & \textbf{53.04} & 20.13 & 1$\times$ \\
    STEM        & 755M & 0.8633 & 0.1483 & 55.72 & \textbf{29.35} & 54.01 & 65.13 & 35.47 & \textbf{34.20} & \underline{52.88} & 20.13 & 1$\times$ \\
    VEmbedding  & 95M  & \underline{0.8620} & 0.1530 & \underline{56.82} & \underline{28.50} & 53.67 & \textbf{66.43} & \textbf{36.23} & 31.20 & 52.09 & \underline{22.21} & 1.03$\times$ \\
    \midrule
    \rowcolor{blue!8}
    \ours-A1/3   & 95M  & 0.8623 & \underline{0.1587} & 56.44 & 27.30 & 50.49 & \underline{66.00} & \underline{35.59} & \underline{34.00} & 52.49 & \textbf{23.31} & 1.07$\times$ \\
    \rowcolor{blue!8}
    \ours-A2/6   & 189M & \textbf{0.8560} & \textbf{0.1686} & \textbf{57.24} & 27.65 & \textbf{57.86} & 65.18 & \textbf{36.19} & 32.00 & 51.93 & 21.98 & 1.07$\times$ \\
    \midrule
    \multicolumn{13}{@{}l}{\textit{Llama/MobileLLM 350M}} \\
    \midrule
    Base & 0M & 0.7750 & 0.1988 & 64.23 & 34.04 & \textbf{55.66} & 70.13 & 47.87 & \textbf{37.20} & 54.14 & 16.68 & 1$\times$ \\
    STEM & 1342M & 0.7697 & 0.2179 & 64.52 & 35.41 & 50.52 & 69.37 & 47.64 & 36.80 & 54.54 & \textbf{25.94} & 1.09$\times$ \\
    VEmbedding & 168M & 0.7651 & \underline{0.2214} & \textbf{66.29} & 36.26 & 47.06 & 70.62 & \textbf{49.60} & \underline{37.00} & \underline{56.35} & 23.87 & 1.04$\times$ \\
%    Engram & 346M & 0.7622 & 0.2207 & 63.26 & 33.70 & 61.74 & 70.57 & 49.08 & 38.80 & 55.17 & 16.52 & 1.08$\times$ \\
    \midrule
    \rowcolor{blue!8}
    \ours-A2/5 & 168M & \underline{0.7647} & 0.2202 & 65.03 & \underline{36.60} & 44.65 & \underline{70.73} & \underline{49.20} & 35.80 & 56.04 & \underline{24.70} & 1.04$\times$ \\
    \rowcolor{blue!8}
    \ours-A2/10 & 336M & \textbf{0.7610} & \textbf{0.2286} & \underline{65.28} & \textbf{36.77} & \underline{55.44} & \textbf{71.60} & \textbf{49.50} & 36.00 & \textbf{56.75} & 21.15 & 1.08$\times$ \\
    \midrule
    \multicolumn{13}{@{}l}{\textit{Qwen3 0.6B}} \\
    \midrule
    Base & 0M & 0.7594 & 0.2491 & 65.99 & 35.32 & \underline{59.88} & 70.78 & 50.58 & 37.60 & 55.96 & 25.72 & 1$\times$ \\
    STEM & 1409M & 0.7564 & \underline{0.2556} & \textbf{67.97} & 36.52 & 56.39 & \textbf{71.98} & 51.01 & \underline{39.00} & \underline{56.59} & 26.01 & 1.07$\times$ \\
    VEmbedding & 470M & \textbf{0.7427} & 0.2530 & 66.50 & \underline{38.05} & 47.55 & 71.00 & \underline{53.07} & \underline{39.00} & \textbf{57.14} & \textbf{26.82} & 1.06$\times$ \\
    \rowcolor{blue!8}
    \ours-A2/8& 470M & \underline{0.7461} & \textbf{0.2737} & \underline{67.09} & \textbf{38.31} & \textbf{63.39} & \underline{71.76} & \textbf{53.39} & \textbf{39.60} & 56.04 & \underline{26.19} & 1.06$\times$ \\
    \bottomrule
  \end{tabular}%
  }
\end{table*}

\textbf{Iso-FLOP results on nanochat architecture.} We compare value embedding, Bigram, and \ours{} on the 12-layer 135M nanochat-style backbone at a fixed $3{\times}10^{18}$ training FLOPs ($\approx$3.3B tokens), inserting one memory block at every odd layer and matching all augmented variants to an iso-151M memory-parameter budget via $N_{\mathrm{hash}}{=}6V$ for Bigram (Table~\ref{tab:pretrain-12layer}). First, at the iso-parameter budget \ours{} improves both validation bpb and CORE over Bigram and VEmbedding; doubling the memory to 302M further lowers validation bpb for all three grouping factors, while CORE improves for $c_{\mathrm{grp}}{=}1$ and $2$. Second, sweeping the row-grouping factor $c_{\mathrm{grp}}\in\{1,2,4\}$ at fixed parameter count, $c_{\mathrm{grp}}{=}2$ gives the highest CORE at both budgets and the lowest train and validation bpb at 302M; at 151M, $c_{\mathrm{grp}}{=}4$ gives slightly lower train and validation bpb. Third, the mixture-of-memory routing adds negligible runtime overhead: \ours{} stays within ${<}2\%$ of the iso-parameter Bigram throughput.

\textbf{Iso-token results on Llama/MobileLLM and Qwen3 architectures.} We compare methods at a fixed token budget within each backbone scale: $5$B tokens for Llama3-style MobileLLM $125$M, and $20$B tokens for MobileLLM $350$M~\citep{grattafiori2024llama3herdmodels,liu2024mobilellmoptimizingsubbillionparameter} and Qwen3 $0.6$B~\citep{yang2025qwen3technicalreport}. Memory-parameter budgets vary by method and configuration. Following the original STEM 1/2 setting~\citep{sadhukhan2026stem}, the memory module is inserted in half of the transformer layers; full per-family architecture details are in Appendix~\ref{sec:appendix-token-matched-arch}. Table~\ref{tab:mobilenet125-summary} supports three conclusions. First, the \ours{} gain over the no-memory base is stable and transfers across both families and across the $125$M, $350$M, and $0.6$B scales. Second, \ours{} improves both validation bpb and CORE over STEM across all three scales while using fewer memory parameters. Third, \ours{} requires $1.04$--$1.08\times$ the training wall time of the no-memory base.

\begin{figure}[!t]
  \noindent\hspace*{-5mm}%
  \begin{minipage}[c]{0.42\linewidth}
    \centering
    \includegraphics[width=\linewidth]{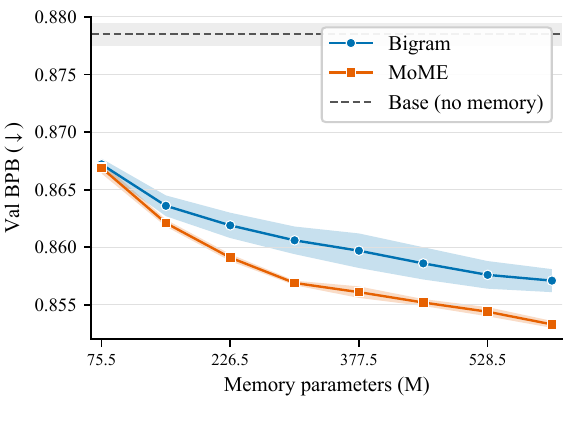}
    \caption{Scaling performance: Bigram vs.\ \ours{}. Colored regions denote error bars.}
    \label{fig:scaling-curve}
  \end{minipage}\hfill
  \begin{minipage}[c]{0.34\linewidth}
    \centering
    % Compound scaling: Engram bigram size N paired with \ours\ M-slot memory
% (K=2 throughout). 12-layer nanochat backbone, 3e18 training FLOPs,
% 3-seed mean. Designed to sit side-by-side with Figure~\ref{fig:scaling-curve}.
\begingroup
\centering
\scriptsize
\setlength{\tabcolsep}{4pt}
\renewcommand{\arraystretch}{1.18}
\begin{tabular}{@{}c r c c@{}}
  \toprule
  $N \times M$ & Mem & Train bpb & Val bpb \\
  \midrule
  Bigram            & 151 & $0.8632$ & $0.8636$ \\
  $6V \times 6$    & 151 & $0.8618$ & $0.8624$ \\
  \midrule
  $12V \times 6$   & 302 & $\mathbf{0.8576}$ & $\mathbf{0.8582}$ \\
  $6V \times 12$   & 302 & $0.8582$ & $0.8588$ \\
  \midrule
  $12V \times 12$  & 604 & $\mathbf{0.8540}$ & $\mathbf{0.8546}$ \\
  $24V \times 6$   & 604 & $0.8541$ & $0.8547$ \\
  $6V \times 24$   & 604 & $0.8544$ & $0.8550$ \\
  \bottomrule
\end{tabular}
\captionof{table}{Compound scaling of Bigram table size $N$ with \ours{} slot count $M$ ($K{=}2$). Bold marks the best values within each memory budget.}
\label{tab:engram-mom-compound-scaling}
\endgroup

  \end{minipage}\hfill
  \begin{minipage}[c]{0.22\linewidth}
    \centering
    % Compact L=2 inference-latency table for the side-by-side layout next to
% Table~\ref{tab:engram-mom-compound-scaling}. Per-step latency overhead
% (percentage increase) at $L_{\mathrm{mem}}{=}2$ relative to the no-memory baseline,
% contrasting our two injection sites. Numbers from Appendix
% Table~\ref{tab:appendix-inference-latency}.
\begingroup
\centering
\scriptsize
\setlength{\tabcolsep}{3pt}
\renewcommand{\arraystretch}{1.18}
\begin{tabular}{@{}l c c@{}}
  \toprule
  Size & \shortstack{Value\\Inject} & \shortstack{Hidden State\\Inject} \\
  \midrule
  \multicolumn{3}{@{}l}{\textit{Llama/MobileLLM}} \\
  $350$M & $7.82\%$ & $8.05\%$ \\
  $1$B   & $5.47\%$ & $5.37\%$ \\
  \midrule
  \multicolumn{3}{@{}l}{\textit{Qwen3}} \\
  $0.6$B & $7.90\%$ & $6.44\%$ \\
  $4$B   & $2.35\%$ & $4.60\%$ \\
  $8$B   & $1.56\%$ & $2.23\%$ \\
  \bottomrule
\end{tabular}
\captionof{table}{Inference-time increase relative to the Base model.}
\label{tab:inference-latency}
\endgroup

  \end{minipage}
\end{figure}

\textbf{Memory-size scaling.}
We further investigate the scaling trend of \ours{} compared with Bigram. We fix the backbone to nanochat d12 and scale the memory parameters under an iso-FLOP setting.
As shown in Figure~\ref{fig:scaling-curve}, \ours{} and Bigram start from similar performance at small memory sizes. \ours{} achieves lower validation bpb than the matched-memory Bigram row at every tested memory size. Within the tested range, scaling \ours{} therefore gives a consistently better bpb trajectory than scaling Bigram.
We also observe that \ours{} has substantially lower run-to-run variance in both training and validation bpb curves (visible as narrower error bands in Figure~\ref{fig:scaling-curve} and per-budget standard deviations in Appendix Table~\ref{tab:appendix-memory-scaling}), suggesting that \ours{} is more stable to train than Bigram in this setting.

\textbf{Compound scaling with Bigram.}
Because \ours{} does not make assumptions about the first-stage token-based indexer $f(x_t)$, it is naturally complementary to methods such as Bigram. We therefore ask whether the two mechanisms can be combined. To test this, we use the baseline's bigram indexer as the first-stage token-based indexer while keeping the overall memory design of \ours{}. We also slightly adjust the configuration so that all memory-augmented layers share the same memory table, enabling an iso-parameter comparison with the Bigram baseline. As shown in Table~\ref{tab:engram-mom-compound-scaling}, combining \ours{} with Bigram improves over Bigram under the same parameter budget. Moreover, at larger budgets, the best-performing variant comes from scaling both the Bigram table and the \ours{} memory slots. This suggests that the two mechanisms are compatible and can be scaled together for further performance gains.

% Queue Table 6 and Figure 4 before the page break so they occupy the top of
% page 9 in that order. Figure 3 is numbered later but intentionally placed at
% the top of page 10, so reserve Figure 4's number here and restore the counter.
\begin{table}[!t]
  \caption{100B-token training results with the nanochat d24 architecture and external base-model references. Network parameters exclude memory parameters; bold marks the best result within each block.}
  \label{tab:d24-100b-result}
  \centering
  \scriptsize
  \setlength{\tabcolsep}{2.4pt}
  \renewcommand{\arraystretch}{1.08}
  \resizebox{\textwidth}{!}{%
  \begin{tabular}{@{}lrrrrccccc@{}}
    \toprule
    {Model} & {Train tokens} & {Network (B)} & {Memory (B)} & {Score tok/s} & {ClimbMix} $\downarrow$ & {FineWeb-Edu} $\downarrow$ & {enwik9} $\downarrow$ & {Shakespeare} $\downarrow$ & {CORE-22} $\uparrow$ \\
    {Qwen3-0.6B} & 36T & 0.60 & 0.00 & 43.8k & 0.7629 & 0.8125 & 0.8649 & 1.3855 & \textbf{0.3753} \\
    {Llama-3.2-1B} & $\leq$9T & 1.24 & 0.00 & 51.7k & \textbf{0.7186} & \textbf{0.7378} & \textbf{0.7451} & \textbf{1.2169} & 0.3629 \\
    \midrule
    {Dense} & 100B & 0.78 & 0.00 & -- & 0.6639 & 0.7677 & 0.9037 & 1.4993 & 0.3441 \\
    {VEmbedding} & 100B & 0.78 & 0.60 & 64.0k & \textbf{0.6463} & 0.7549 & 0.8860 & 1.4823 & 0.3484 \\
    \rowcolor{blue!8}
    {\ours{}-A2/12} & 100B & 0.78 & 0.61 & 58.4k & 0.6504 & \textbf{0.7548} & \textbf{0.8850} & \textbf{1.4789} & \textbf{0.3687} \\
    \bottomrule
  \end{tabular}%
  }
\end{table}

\setcounter{figure}{3}
\begin{figure}[!t]
  \centering
  \begin{minipage}[t]{0.248\textwidth}
    \vspace{0pt}
    \centering
    \includegraphics[width=\linewidth]{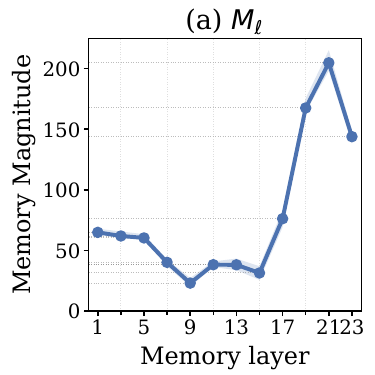}
  \end{minipage}%
  \begin{minipage}[t]{0.248\textwidth}
    \vspace{0pt}
    \centering
    \includegraphics[width=\linewidth]{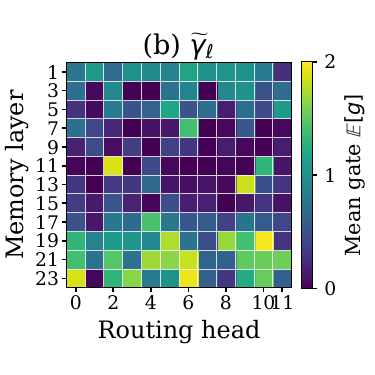}
  \end{minipage}%
  \begin{minipage}[t]{0.248\textwidth}
    \vspace{0pt}
    \centering
    \includegraphics[width=\linewidth]{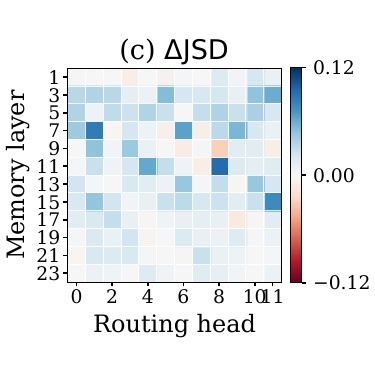}
  \end{minipage}%
  \begin{minipage}[t]{0.248\textwidth}
    \vspace{0pt}
    \centering
    \includegraphics[width=\linewidth]{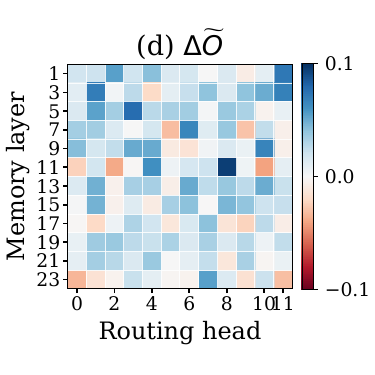}
  \end{minipage}
  \caption{Model analysis. (a) CORE injected-memory norm by layer. (b) Mean CORE injection gate. (c) WiC routing-distribution effect. (d) WiC $\Delta\widetilde O$ effect. Blue indicates the hypothesis-consistent direction.}
  \label{fig:main-model-analysis}
\end{figure}
\setcounter{figure}{2}

\textbf{Inference latency.}
Although we report training FLOPs and end-to-end wall time to evaluate the efficiency of \ours{}, we further probe the inference-latency overhead introduced by the memory mixture as the backbone scales. As shown in Table~\ref{tab:inference-latency}, we measure per-step inference latency for a memory-augmented transformer block and report the percentage increase relative to the Base model. We find that the routing and memory-retrieval cost is small compared with the rest of the transformer computation, so the relative latency overhead decreases as the backbone becomes larger. We also ablate the injection design: in \ours{}, the router takes the hidden state as input and injects the retrieved memory into the value stream, making routing and retrieval a parallel path to the standard value projection. Compared with a hidden-state-to-hidden-state injection path for \ours{}, this design shows a clear latency advantage as the model scales.

\textbf{Scaled-up training on ClimbMix.}
We also evaluate \ours{} in a scaled-up training setting.
We compare \ours{}, Dense, and VEmbedding after training on 104.858B ClimbMix tokens~\citep{diao2025nemotronclimb} in Table~\ref{tab:d24-100b-result}. We also include the Qwen3-0.6B~\citep{yang2025qwen3technicalreport} and Llama~3.2-1B~\citep{meta2024llama32modelcard} base models as external references trained on substantially larger corpora. We report bits per byte (bpb) on four corpora: the held-out ClimbMix validation set is in-domain, while FineWeb-Edu~\citep{lozhkov2024finewebedu}, enwik9, and Shakespeare serve as out-of-domain evaluation sets. 
Within the controlled 100B-token block of Table~\ref{tab:d24-100b-result}, \ours{} obtains lower bpb than Dense and VEmbedding on all three out-of-domain evaluations and a higher CORE-22 score, while its in-domain ClimbMix bpb is slightly higher than that of VEmbedding. \ours{} also approaches Qwen3-0.6B on CORE-22 despite using substantially fewer pretraining tokens, and its measured full-sequence scoring throughput is in the same range as both external references.

\subsection{\texorpdfstring{Model Analysis}{Model Analysis}}\label{sec:model-analysis}

\textbf{Memory usage per layer.} To measure memory use across depth, we compute a task-balanced average over token positions and value heads at each memory layer $\ell$ with the model reported in Table~\ref{tab:d24-100b-result}. We summarize the per-head injection gate and the magnitude of the final memory residual added to the value head as
\begin{equation}
M_{\ell}
=\mathbb{E}_{\substack{(x,t)\sim\mathrm{CORE}\\ i\in[H]}}
  \!\left[\left\|\gamma_{\ell,t,i}
  \tilde{\mathbf{m}}_{\ell,t,i}\right\|_2\right],
\qquad
\widetilde{\gamma}_{\ell}
=\mathbb{E}_{\substack{(x,t)\sim\mathrm{CORE}\\ i\in[H]}}
  [\gamma_{\ell,t,i}],
\label{eq:core-layer-memory-use}
\end{equation}
where $\gamma_{\ell,t,i}\tilde{\mathbf{m}}_{\ell,t,i}$ is the gated memory vector ultimately infused into value head $i$. Panel~(a) of Figure~\ref{fig:main-model-analysis} plots $M_{\ell}$, while panel~(b) plots $\widetilde{\gamma}_{\ell}$. Both measurements show the same depth profile: memory use is strongest near the beginning and especially the end of the model, with substantially lower utilization through the middle layers. This pattern is consistent with memory contributing to early contextualization and then being re-engaged for late task-specific refinement, while the middle layers rely more heavily on the transformed backbone representations. Appendix~\ref{sec:appendix-memory-domain-audit} provides the per-head and per-domain measurements.

\input{tables/main_semantic_routes}
\setcounter{figure}{4}

\textbf{Semantic interpretability in \ours{}.}
One initial motivation for \ours{} is to capture contextual variation in token semantics. This semantic structure does appear in the learned routing. We probe the learned router on a set of polysemous tokens and find that same-sense prompts tend to route to the same memory slot.
Figure~\ref{fig:main-semantic-routes} visualizes this behavior for three polysemous tokens (\emph{bank}, \emph{bug}, and \emph{drive}) at a single Qwen-0.6B head. Same-sense prompts route through the same memory slot, while the changed-sense prompt jumps to a different slot. More results are provided in Appendix~\ref{sec:appendix-semantic-routes}.

\textbf{Quantitative analysis of semantic routing.}
Going beyond qualitative analysis, we quantify sense-sensitive routing using the Word-in-Context (WiC) dataset~\citep{pilehvar2019wic}, which pairs two natural sentences containing the same target word and labels whether its sense is the same or different. We evaluate 670 strictly filtered pairs at all 144 memory layer--head sites in the d24 checkpoint.
We use two complementary metrics to test a simple expectation: if routing reflects word sense, it should change more between different-sense pairs than between same-sense pairs. Jensen--Shannon divergence (JSD) compares the full routing distributions, capturing differences in weights even when the selected slots stay the same. The chance-corrected top-2 decision overlap $\widetilde O$ of \citet{olson-etal-2025-probing} measures how much the two occurrences select the same slots. Using both checks whether the pattern holds for both routing weights and discrete slot choices. At each site, we report the mean JSD for different-sense pairs minus that for same-sense pairs ($\Delta\mathrm{JSD}$), and the mean overlap with the order reversed ($\Delta\widetilde O$). Positive values in either measure therefore indicate more similar routing for the same sense than for different senses. Panels~(c)--(d) of Figure~\ref{fig:main-model-analysis} show positive $\Delta\mathrm{JSD}$ at 117 of 144 sites and positive $\Delta\widetilde O$ at 110 of 144 sites. Thus, most memory-routing heads exhibit similar routing for the same semantic context and different routing when the context changes the sense. Appendix~\ref{sec:appendix-wic-routing} provides the full filtering protocol, metric definitions, per-head statistics, gate-aware sensitivity analysis, and limitations.

\FloatBarrier

% Discussion
\section{Discussion}\label{sec:discussion}
\textbf{Limitations.} The main limitation of this work is computational scale. Although the extended 100B-token experiment reaches approximately $1.4$B total parameters, the underlying dense backbone has approximately $0.8$B parameters and this scale check uses a single seed. The CORE activation and WiC analyses provide descriptive evidence of selective memory use and sense-sensitive routing, but causal interventions are needed to establish whether these behaviors improve downstream accuracy or robustness under domain shift.

\textbf{Future work.}
Future work should investigate better compound-scaling strategies that combine efficient token routing with the mixture-of-memory mechanism. A second direction is suggested by the semantic structure in the learned routing: whether controlling memory routes can provide a practical handle for steering model behavior.

\textbf{Conclusion.}
We presented \ours{}, a conditional memory mechanism that keeps efficient token-indexed access while allowing context-dependent selection among multiple memory slots. Across controlled pretraining experiments on nanochat, Llama/MobileLLM, and Qwen3-style backbones, \ours{} improves over strong memory baselines and shows favorable scaling and latency behavior in the tested regimes. These results suggest that context-aware memory mixtures are a practical direction for expanding sparse model capacity.

\bibliographystyle{plainnat}
\addtolength{\bibsep}{-1pt}
\bibliography{custom}

\clearpage
\appendix
\section{Appendix}

\subsection{\texorpdfstring{Notation}{Notation}}
\label{sec:appendix-notation}

\begin{table*}[!htbp]
  \caption{Notation used for Mixture of Memory. We reserve $V$ for vocabulary size, $N$ for first-stage memory rows, $M$ for slots per row, $K$ for activated slots, and $d_{\mathrm{value}}$ for the per-head value dimension.}
  \label{tab:notation}
  \centering
  \small
  \setlength{\tabcolsep}{4pt}
  \renewcommand{\arraystretch}{1.15}
  \begin{tabular}{@{}p{0.20\textwidth}p{0.25\textwidth}p{0.47\textwidth}@{}}
    \toprule
    Object & Symbol & Meaning \\
    \midrule
    Index shorthand & $[q]$ & The set $\{1,\dots,q\}$. \\
    Input token & $x_t \in \mathcal{V}$ & Token at sequence position $t$. \\
    Vocabulary size & $V=\lvert\mathcal{V}\rvert$ & Number of tokens in the vocabulary. \\
    Hidden state & $\mathbf{h}_t \in \mathbb{R}^{d_{\mathrm{model}}}$ & Transformer hidden state used by the slot gate and value-residual gate. \\
    Value heads & $H$ & Number of key/value heads at the memory-injection layer. \\
    Value dimension & $d_{\mathrm{value}}$ & Per-head value dimension; $\mathbf{v}_{t,i}\in\mathbb{R}^{d_{\mathrm{value}}}$ and $\mathbf{v}_t\in\mathbb{R}^{H d_{\mathrm{value}}}$. \\
    Augmented value & $\tilde{\mathbf{v}}_{t,i}\in\mathbb{R}^{d_{\mathrm{value}}}$ & Value vector after adding the gated retrieved memory for position $t$ and value head $i$. \\
    Row indexer & $f:\mathcal{V}\to[N]$, $n_t=f(x_t)$ & Fixed first-stage map from token id to memory row. \\
    Slot gate & $g_\theta^{(i)}(\mathbf{h}_t)$ & Second-stage context indexer returning the activated slot set $\mathcal{A}_{t,i}$ and mixture weights $\alpha^{(i)}_{t,a}$ for head $i$. \\
    Grouping factor & $c_{\mathrm{grp}}$ & Target token grouping/compression factor for the kNN slot map; this is distinct from the active-slot count $K$. \\
    Memory bank & $\mathbf{E}^{\mathrm{mem}}\in\mathbb{R}^{N\times M\times d_{\mathrm{value}}}$ & Learnable memory tensor with $M$ candidate slots per row, shared across value heads. \\
    Stored memory vector & $\mathbf{m}_{n,a}=\mathbf{E}^{\mathrm{mem}}[n,a,:]$ & Memory vector stored at row $n$ and slot $a$, with $\mathbf{m}_{n,a}\in\mathbb{R}^{d_{\mathrm{value}}}$. \\
    Gate logits & $\boldsymbol{\ell}_{t,i}\in\mathbb{R}^{M}$ & Pre-sigmoid slot logits for position $t$ and value head $i$. \\
    Gate scores & $\mathbf{s}_{t,i}=\sigma(\boldsymbol{\ell}_{t,i})$ & Post-sigmoid slot scores used for top-$K$ selection. \\
    Activated slots & $\mathcal{A}_{t,i}\subseteq[M]$ & Top-$K$ memory slots activated for position $t$ and value head $i$. \\
    Mixture weights & $\alpha^{(i)}_{t,a}$ & Sigmoid-norm weight for slot $a$ after normalization over $\mathcal{A}_{t,i}$. \\
    Retrieved memory & $\tilde{\mathbf{m}}_{t,i}\in\mathbb{R}^{d_{\mathrm{value}}}$ & Per-head memory vector added to the value stream after scaling by $\gamma_{t,i}$. \\
    Value-residual gate & $\boldsymbol{\gamma}_t\in\mathbb{R}^{H}$ & Per-head scalar gate produced from $\mathbf{h}_t$ to control memory injection strength. \\
    \bottomrule
  \end{tabular}
\end{table*}

\FloatBarrier

\subsection{\texorpdfstring{Evaluation Protocol}{Evaluation Protocol}}
\label{sec:appendix-eval-protocol}

Table~\ref{tab:appendix-evaluation-suite} lists the task suite used by the
local nanochat CORE evaluator. The evaluator shuffles each task with a fixed
seed, evaluates raw accuracy, reads the random baseline for each task from the
evaluation metadata, and reports the unweighted mean of centered accuracies as
CORE. For bpb, the evaluator sums target negative log-likelihood over counted
tokens, masks ignored and special-token targets, converts nats to bits, and
normalizes by the UTF-8 byte count of the same targets.

\begin{table*}[!ht]
  \caption{The 22-task CORE evaluation suite used by the local nanochat evaluation bundle. The evaluator reports raw task accuracy and then averages task-specific random-baseline-centered accuracies for the CORE metric.}
  \label{tab:appendix-evaluation-suite}
  \centering
  \scriptsize
  \setlength{\tabcolsep}{4pt}
  \renewcommand{\arraystretch}{1.12}
  \begin{tabular}{@{}l c l l@{}}
    \toprule
    Task & Shots & Evaluation type & Source \\
    \midrule
    HellaSwag-0 & 0 & multiple choice & \citep{hellaswag} \\
    Jeopardy & 10 & language modeling & \citep{kaggle200000Jeopardy} \\
    BB-QA Wikidata & 10 & language modeling & \citep{srivastava2023beyond} \\
    ARC-Easy & 10 & multiple choice & \citep{arc} \\
    ARC-Challenge & 10 & multiple choice & \citep{arc} \\
    COPA & 0 & multiple choice & \citep{copa} \\
    CommonsenseQA & 10 & multiple choice & \citep{talmor-etal-2019-commonsenseqa} \\
    PIQA & 10 & multiple choice & \citep{piqa} \\
    OpenBookQA & 0 & multiple choice & \citep{OpenBookQA2018} \\
    LAMBADA OpenAI & 0 & language modeling & \citep{lambada} \\
    HellaSwag & 10 & multiple choice & \citep{hellaswag} \\
    Winograd & 0 & schema & \citep{winograd} \\
    WinoGrande & 0 & schema & \citep{sakaguchi2019winogrande} \\
    BB-Dyck Languages & 10 & language modeling & \citep{srivastava2023beyond} \\
    AGI Eval LSAT-AR & 3 & multiple choice & \citep{zhong2023agieval} \\
    BB-CS Algorithms & 10 & language modeling & \citep{srivastava2023beyond} \\
    BB-Operators & 10 & language modeling & \citep{srivastava2023beyond} \\
    BB-Repeat Copy Logic & 10 & language modeling & \citep{srivastava2023beyond} \\
    SQuAD & 10 & language modeling & \citep{squad} \\
    CoQA & 0 & language modeling & \citep{reddy-etal-2019-coqa} \\
    BoolQ & 10 & multiple choice & \citep{boolq} \\
    BB-Language Identification & 10 & multiple choice & \citep{srivastava2023beyond} \\
    \bottomrule
  \end{tabular}
\end{table*}

\FloatBarrier

\subsection{\texorpdfstring{Training Setup}{Training Setup}}
\label{sec:appendix-training-setup}

Table~\ref{tab:appendix-training-setup} summarizes the training settings. Tables~\ref{tab:appendix-architecture-families} and~\ref{tab:appendix-mome-family-configs} give the backbone and memory configurations for each family.

\begin{table*}[!ht]
  \caption{Training setup for the experiments in Section~\ref{sec:results-nanochat}. Architecture, memory configuration, device batch size, and training budget vary by comparison.}
  \label{tab:appendix-training-setup}
  \centering
  \scriptsize
  \setlength{\tabcolsep}{4pt}
  \renewcommand{\arraystretch}{1.15}
  \begin{tabular}{@{}p{0.20\textwidth} p{0.72\textwidth}@{}}
    \toprule
    Item & Setting \\
    \midrule
    Data & The experiments in Tables~\ref{tab:pretrain-12layer}--\ref{tab:engram-mom-compound-scaling} use FineWeb-Edu~\citep{lozhkov2024finewebedu}. Our 100B-token runs in Table~\ref{tab:d24-100b-result} and the d24 appendix controls use ClimbMix~\citep{diao2025nemotronclimb}. \\
    Tokenizer & Shared byte-level BPE tokenizer~\citep{radford2019language} with vocabulary size $32{,}768$, trained on the FineWeb-Edu training stream; special tokens have zero byte count for bpb. \\
    Context length & $2048$ \\
    Total batch & Tables~\ref{tab:pretrain-12layer}--\ref{tab:engram-mom-compound-scaling} use $524{,}288$ tokens; our 100B-token runs in Table~\ref{tab:d24-100b-result} and the d24 appendix controls use $1{,}048{,}576$ tokens. Gradient accumulation preserves the total batch size as device batch size and world size vary. \\
    Optimizer groups & Muon is used for matrix-shaped transformer parameters; AdamW groups are used for embeddings, unembeddings, scalar parameters, value-memory tables, and other non-matrix parameters. Value-memory tables may use a separate AdamW learning rate and weight decay, listed in Table~\ref{tab:appendix-mome-family-configs}. \\
    Learning-rate schedule & Run launchers use a warmup followed by cosine decay or warmdown. The nanochat scaling runs use target-FLOP stopping; token-matched Llama/MobileLLM and Qwen3-style runs use fixed token budgets. \\
    Evaluation tokens & BPB evaluation uses the run-specified token budget: $20{,}971{,}520$ tokens per split for nanochat/scaling reports and $2{,}097{,}152$ tokens per split for token-matched Llama/MobileLLM and Qwen3 reports. CORE uses all examples unless a run explicitly sets a per-task cap. \\
    Seed policy & Nanochat scaling tables report three-seed means over seeds $42,43,44$ unless otherwise noted. Token-matched family tables use the completed run(s) available for each configuration. \\
    Distributed training & Runs use DDP. The local reports use world size $4$ for most nanochat and Llama/MobileLLM runs and world size $8$ for Qwen3-style and larger nanochat d24 runs. \\
    Hardware & \leavevmode{}Training hardware varies by run, including NVIDIA RTX PRO 6000 Blackwell (96GB) and H100 (80GB) GPUs. The d24 control reruns use four H100 GPUs; the sparse-MoE throughput comparison in Table~\ref{tab:appendix-sparse-moe-interaction} uses eight H100 GPUs. \\
    \bottomrule
  \end{tabular}
\end{table*}

\FloatBarrier

\subsection{\texorpdfstring{Backbone Family Details}{Backbone Family Details}}
\label{sec:appendix-token-matched-arch}

Table~\ref{tab:appendix-architecture-families} lists the backbone configurations for the nanochat, Llama/MobileLLM, and Qwen3-style experiments.

\begin{table*}[!ht]
  \caption{Backbone configurations for the reported experiments. Asterisks mark dimensions derived from the configured aspect ratio and head dimension.}
  \label{tab:appendix-architecture-families}
  \centering
  \scriptsize
  \setlength{\tabcolsep}{3pt}
  \renewcommand{\arraystretch}{1.15}
  \resizebox{\textwidth}{!}{%
  \begin{tabular}{@{}l l l r r r r r c r l@{}}
    \toprule
    Family & Scale & Model type & Depth & $d_{\mathrm{model}}$ & Head dim & Attn. heads & KV heads & Tie emb. & Ctx. & Training budget \\
    \midrule
    nanochat & d12 & \texttt{gpt} & 12 & 768* & 128 & 6* & 6* & No & 2048 & $3{\times}10^{18}$ FLOPs; $3.3$B tokens \\
    nanochat & d24 & \texttt{gpt} & 24 & 1536* & 128 & 12* & 12* & No & 2048 & $6{\times}10^{19}$ FLOPs; $12$B tokens \\
    Llama/MobileLLM & 125M & \texttt{stemgpt\_350m} & 30 & 576 & 64 & 9* & 3 & Yes & 2048 & 5B tokens \\
    Llama/MobileLLM & 350M & \texttt{stemgpt\_350m} & 32 & 960 & 64 & 15* & 5 & No & 2048 & 20B tokens \\
    Qwen3-style & 0.6B & \texttt{qwen3\_0p5b} & 28 & 1024 & 128 & 16 & 8 & Yes & 2048 & 20B tokens \\
    \bottomrule
  \end{tabular}%
  }
\end{table*}

\FloatBarrier

\subsection{\texorpdfstring{Memory Module Configurations Across Families}{Memory Module Configurations Across Families}}
\label{sec:appendix-mome-family-configs}

Table~\ref{tab:appendix-mome-family-configs} specifies the memory layers, row indexer, router input, gate function, and value-table optimizer for each backbone family.

\begin{table*}[!ht]
  \caption{Memory configurations by backbone family. A$K$/$M$ means that the router selects $K$ of $M$ slots per row. ``Launcher-default wd'' uses the default weight decay for the value-memory table in that run.}
  \label{tab:appendix-mome-family-configs}
  \centering
  \small
  \begin{minipage}{\linewidth}
  \begin{description}
    \setlength{\itemsep}{4pt}
    \setlength{\parskip}{0pt}

    \item[\textbf{nanochat d12}.]
    Memory layers: memory blocks with span $1$.\;
    Shape: $N\times M\times d_{\mathrm{value}}$.\;
    \ours{} setting: A2/6, A2/12, A2/24 in main rows, with $M$ swept in the
    scaling study.\;
    Row indexer $f$: identity or frozen kNN table, depending on row.\;
    Router input: hidden.\;
    Gate: sigmoid-norm.\;
    Value-table optimizer: AdamW group, lr $0.2$, wd $0.001$.

    \item[\textbf{nanochat d24}.]
    Memory layers: odd layers $1,3,\ldots,23$.\;
    Shape: $N\times M\times d_{\mathrm{value}}$.\;
    \ours{} setting: A2/12.\;
    Row indexer $f$: identity.\;
    Router input: hidden.\;
    Gate: sigmoid-norm.\;
    Value-table optimizer: AdamW group, lr $0.2$, wd $0.001$.

    \item[\textbf{Llama/MobileLLM 125M}.]
    Memory layers: odd layers $1,3,\ldots,29$.\;
    Shape: $V\times M\times d_{\mathrm{value}}$.\;
    \ours{} setting: A1/3 and A2/6 reported variants.\;
    Row indexer $f$: identity.\;
    Router input: value.\;
    Gate: softmax for A1/3, sigmoid-norm for A2/6.\;
    Value-table optimizer: AdamW group, lr $0.1$, wd $0.001$.

    \item[\textbf{Llama/MobileLLM 350M}.]
    Memory layers: odd layers $1,3,\ldots,31$.\;
    Shape: $V\times M\times d_{\mathrm{value}}$.\;
    \ours{} setting: A2/5 and A2/10.\;
    Row indexer $f$: identity.\;
    Router input: value.\;
    Gate: sigmoid-norm.\;
    Value-table optimizer: AdamW group, lr $0.1$, launcher-default wd.

    \item[\textbf{Qwen3-style 0.6B}.]
    Memory layers: odd layers $1,3,\ldots,27$.\;
    Shape: $V\times M\times d_{\mathrm{value}}$.\;
    \ours{} setting: A2/8.\;
    Row indexer $f$: identity.\;
    Router input: hidden.\;
    Gate: sigmoid-norm.\;
    Value-table optimizer: AdamW group, lr $0.2$, wd $0.001$.
  \end{description}
  \end{minipage}
\end{table*}

\FloatBarrier

\subsection{\texorpdfstring{Baseline Implementation Details}{Baseline Implementation Details}}
\label{sec:appendix-baseline-implementation}

\paragraph{Bigram.}
Our Bigram baseline follows modded-nanogpt practice~\citep{classiclarry2026bigramhash}, rather than the canonical Engram architecture of \citet{cheng2026conditional}. Memory is a single table of $N$
rows with row dimension $d_{\mathrm{model}}$, indexed at each position by a
deterministic hash of the previous token together with the current token (the
bigram key). The same table is shared across the memory-augmented layers, which are the odd-indexed blocks in the default nanochat configuration. At each such layer, the retrieved row is added to the residual stream through hidden-state-conditioned per-head gates and a learned per-layer scalar. The lookup itself remains a deterministic function of the two token IDs. The dictionary size is written as a multiplier of the
tokenizer vocabulary size $V$; memory-scaling sweeps vary this multiplier
while keeping depth, target FLOPs, context length, batch size, and seed
policy aligned with the matched \ours{} runs. The Mobile/Llama Bigram row
uses the matched-size 350M run when it is reported.

At approximately matched total parameters and $6\times10^{19}$ training FLOPs on d24 ClimbMix, Bigram achieves lower validation bpb and higher CORE-22 than our canonical Engram adaptation (Table~\ref{tab:appendix-token-memory-adaptations}). This single-seed comparison uses an adaptation that retains 2-/3-gram lookup and hidden-state-conditioned fusion but omits Engram's multi-stream integration for the single-stream nanochat backbone.

\paragraph{Value Embedding.}
The Value Embedding (VE) baseline follows modded-nanogpt and nanochat~\citep{koszarsky2024valueembeddings,karpathy2025nanochat} and the value-residual formulation of \citet{zhou2024value}. A token-indexed table of size $V\times H\times d_{\mathrm{value}}$ supplies one embedding slice per attention head, which is added to the attention value stream. By default, VE is injected into the odd-indexed transformer blocks; its table size is set by the vocabulary and value-head dimensions.

\paragraph{STEM.}
The STEM baseline follows the original architecture construction and
released implementation~\citep{sadhukhan2026stem}. We adapt the reported
one-half-layer STEM setting to the same training harness, tokenizer, context
length, and batch size used by the Llama/MobileLLM-family comparisons.

\subsection{\texorpdfstring{Compound Scaling}{Compound Scaling}}
\label{sec:appendix-compound-scaling}

\subsubsection{\texorpdfstring{Bigram Memory and Slot Scaling}{Bigram Memory and Slot Scaling}}

Table~\ref{tab:appendix-compound-scaling} expands Table~\ref{tab:engram-mom-compound-scaling} with total parameters and CORE. Each fixed-budget group compares allocations to Bigram table size, \ours{} slot count, or both.

\begin{table}[!ht]
  \caption{Full compound scaling table for Bigram table size $N$ and \ours{} slot count $M$ on the 12-layer nanochat backbone, averaged over three seeds. The compact main table reports only memory size and bpb; this table additionally lists total parameters and CORE. Bold and underline mark the best and second-best train/validation bpb within each fixed memory-budget group.}
  \label{tab:appendix-compound-scaling}
  \centering
  \scriptsize
  \setlength{\tabcolsep}{4pt}
  \renewcommand{\arraystretch}{1.15}
  \begin{tabular}{@{}c c r r c c c@{}}
    \toprule
    Bigram $N$ & \ours{} setting & Mem (M) & Total params (M) & Train bpb $\downarrow$ & Val bpb $\downarrow$ & CORE $\uparrow$ \\
    \midrule
    $6V$  & A2/6  & 151 & 286.4 & \metricstd{0.8618}{0.0010} & \metricstd{0.8624}{0.0010} & \metricstd{0.1647}{0.0023} \\
    \midrule
    $12V$ & A2/6  & 302 & 437.4 & \bestmetricstd{0.8576}{0.0008} & \bestmetricstd{0.8582}{0.0007} & \metricstd{0.1549}{0.0084} \\
    $8V$  & A2/9  & 302 & 437.5 & \secondmetricstd{0.8579}{0.0010} & \secondmetricstd{0.8585}{0.0010} & \metricstd{0.1551}{0.0088} \\
    $6V$  & A2/12 & 302 & 437.6 & \metricstd{0.8582}{0.0008} & \metricstd{0.8588}{0.0008} & \metricstd{0.1544}{0.0095} \\
    \midrule
    $12V$ & A2/12 & 604 & 739.6 & \bestmetricstd{0.8540}{0.0002} & \bestmetricstd{0.8546}{0.0001} & \metricstd{0.1604}{0.0050} \\
    $24V$ & A2/6  & 604 & 739.4 & \secondmetricstd{0.8541}{0.0011} & \secondmetricstd{0.8547}{0.0012} & \metricstd{0.1590}{0.0096} \\
    $6V$  & A2/24 & 604 & 739.9 & \metricstd{0.8544}{0.0012} & \metricstd{0.8550}{0.0012} & \metricstd{0.1614}{0.0108} \\
    \bottomrule
  \end{tabular}
\end{table}

\FloatBarrier

\subsubsection{\texorpdfstring{Combination with Sparse-FFN MoE}{Combination with Sparse-FFN MoE}}
\label{sec:appendix-sparse-moe-interaction}

Table~\ref{tab:appendix-sparse-moe-interaction} tests whether \ours{} can complement a sparse-FFN MoE. The matched-total E6/K2 MoE-FFN obtains the lowest validation bpb, while combining it with \ours{} gives the highest CORE-22. The combined model has a similar active-parameter count but more stored parameters than either component, and trains more slowly.

\begin{table*}[!htbp]
  \caption{Sparse-MoE interaction on d24/w1536/h12, ClimbMix, seed~42, at approximately $6\times10^{19}$ training FLOPs. E6/K2 activates two of six routed FFN experts in every Transformer layer and retains one always-on shared expert; ``+ \ours{}'' additionally places MoME at every odd layer. Throughput is measured on eight H100 GPUs with the same global batch and native Flash Attention~3. The models have similar active-parameter counts, but the combined model has more stored parameters than either component. Active counts include full input/output embedding matrices, routers, other dense/shared parameters, selected FFN experts, and distinct retrieved memory entries; unselected FFN experts and unaccessed memory entries are excluded.}
  \label{tab:appendix-sparse-moe-interaction}
  \centering
  \small
  \setlength{\tabcolsep}{7pt}
  \renewcommand{\arraystretch}{1.12}
  \resizebox{\textwidth}{!}{%
  \begin{tabular}{@{}lrrrrr@{}}
    \toprule
    Model & Stored parameters (M) & Active parameters/token (M) & Training tokens/s & Validation bpb $\downarrow$ & CORE-22 $\uparrow$ \\
    \midrule
    \rowcolor{blue!8}
    \ours{} & 1,386.8 & 782.8 & \textbf{796,258} & 0.6990 & 0.2941 \\
    MoE-FFN, E6/K2/24 & 1,384.3 & 780.4 & 653,033 & \textbf{0.6903} & 0.2886 \\
    MoE-FFN, E6/K2/24 + \ours{} & 1,991.0 & 783.0 & 611,249 & 0.6913 & \textbf{0.2986} \\
    \bottomrule
  \end{tabular}%
  }
\end{table*}

\FloatBarrier

\subsection{\texorpdfstring{Supplementary Quantitative Results}{Supplementary Quantitative Results}}
\label{sec:appendix-supplementary-quantitative}

\subsubsection{\texorpdfstring{Matched 100B checkpoints and external base-model references}{Matched 100B checkpoints and external base-model references}}
\label{sec:appendix-external-base-models}

Table~\ref{tab:appendix-external-core-full} reports the full 22-task accuracies for the matched 100B dense value-embedding and \ours{} checkpoints, alongside Qwen and Llama references up to the 3B model class. The native checkpoints form a controlled comparison; the external models differ in training data, token exposure, and tokenizer and serve as reference points.

\begin{table*}[!htbp]
  \caption{Raw accuracy (\%) on the 22 CORE tasks for the matched 100B dense value-embedding and \ours{} checkpoints and external Qwen and Llama references up to the 3B model class. Appendix~\ref{sec:appendix-eval-protocol} defines CORE aggregation.}
  \label{tab:appendix-external-core-full}
  \centering
  \scriptsize
  \setlength{\tabcolsep}{4pt}
  \renewcommand{\arraystretch}{1.04}
  \resizebox{\textwidth}{!}{%
  \begin{tabular}{@{}lrrrrrr@{}}
    \toprule
    CORE task & Dense VE & \ours{} & Qwen3-0.6B & Qwen3-1.7B & Llama-3.2-1B & Llama-3.2-3B \\
    \midrule
    HellaSwag, zero-shot & 66.36 & 66.75 & 52.13 & 64.86 & 63.59 & 73.43 \\
    Jeopardy & 26.12 & 28.34 & 25.22 & 40.58 & 34.91 & 48.18 \\
    BigBench QA Wikidata & 56.44 & 57.27 & 59.93 & 68.81 & 69.34 & 72.15 \\
    ARC-Easy & 74.96 & 74.24 & 73.19 & 78.87 & 68.69 & 75.34 \\
    ARC-Challenge & 47.27 & 46.50 & 43.69 & 54.44 & 38.14 & 48.04 \\
    COPA & 69.00 & 70.00 & 71.00 & 75.00 & 74.00 & 80.00 \\
    CommonsenseQA & 50.12 & 59.38 & 66.01 & 80.10 & 36.77 & 71.42 \\
    PIQA & 78.07 & 78.18 & 71.16 & 76.12 & 75.52 & 79.11 \\
    OpenBookQA & 45.20 & 43.20 & 34.80 & 41.80 & 38.80 & 43.80 \\
    LAMBADA & 52.78 & 52.49 & 54.59 & 62.51 & 62.16 & 69.57 \\
    HellaSwag & 67.38 & 67.97 & 52.45 & 65.55 & 65.16 & 75.49 \\
    Winograd & 80.95 & 80.22 & 75.46 & 80.59 & 82.42 & 85.71 \\
    WinoGrande & 61.80 & 60.46 & 58.25 & 63.77 & 60.22 & 69.53 \\
    BigBench Dyck Languages & 4.40 & 7.80 & 12.20 & 31.50 & 13.70 & 22.40 \\
    AGI Eval LSAT-AR & 24.78 & 25.22 & 24.35 & 26.09 & 23.91 & 24.35 \\
    BigBench CS Algorithms & 37.35 & 43.64 & 47.73 & 73.79 & 46.21 & 62.20 \\
    BigBench Operators & 18.57 & 23.33 & 60.95 & 73.33 & 40.00 & 58.10 \\
    BigBench Repeat Copy Logic & 3.12 & 3.12 & 15.62 & 18.75 & 9.38 & 21.88 \\
    SQuAD & 52.12 & 52.53 & 57.75 & 65.51 & 51.97 & 61.00 \\
    CoQA & 38.13 & 39.30 & 39.45 & 44.91 & 36.34 & 43.76 \\
    BoolQ & 62.51 & 69.36 & 74.25 & 81.28 & 64.98 & 74.28 \\
    BigBench Language Identification & 25.42 & 26.31 & 36.58 & 42.71 & 24.77 & 49.00 \\
    \bottomrule
  \end{tabular}%
  }
\end{table*}

\FloatBarrier

\subsubsection{\texorpdfstring{Combined full token-matched results}{Combined full token-matched results}}
\label{sec:appendix-token-matched-full}

Table~\ref{tab:appendix-token-matched-full} combines the Llama/MobileLLM
and Qwen3-style rows from Table~\ref{tab:mobilenet125-summary}. It adds train
bpb and reports raw accuracy for every benchmark in the evaluation suite,
including all BigBench tasks as separate rows.

% Generated by scripts/token_matched_full_table.py.

\begin{table*}[p]
  \caption{Transposed full token-matched results for the runs summarized in Table~\ref{tab:mobilenet125-summary}. Columns are runs grouped by backbone family; rows report memory, relative wall time, train/validation bpb, CORE, and raw benchmark accuracy (\%) for every benchmark in the evaluation suite. All MobileLLM 125M columns report seed 42 from the September 2026 reproduction. BigBench rows are QA Wikidata (QAW), Dyck Languages (Dyck), CS Algorithms (CS), Operators (Op), Repeat Copy Logic (RCL), and Language Identification (LI). Bold marks the best value and underline marks the second-best value within each family block; accuracy values within 0.1 percentage points share the same rank.}
  \label{tab:appendix-token-matched-full}
  \centering
  \scriptsize
  \setlength{\tabcolsep}{2pt}
  \renewcommand{\arraystretch}{1.08}
  \resizebox{\textwidth}{!}{%
  \begin{tabular}{@{}l | ccccc | ccccc | cccc@{}}
    \toprule
     & \multicolumn{5}{c}{Llama/MobileLLM 125M/5B} & \multicolumn{5}{c}{Llama/MobileLLM 350M/20B} & \multicolumn{4}{c}{Qwen3 0.6B/20B} \\
    \cmidrule(lr){2-6} \cmidrule(lr){7-11} \cmidrule(lr){12-15}
    Metric & Base & STEM & VEmbedding & \cellcolor{blue!8}\ours-A1/3 & \cellcolor{blue!8}\ours-A2/6 & Base & STEM & VEmbedding & \cellcolor{blue!8}\ours-A2/5 & \cellcolor{blue!8}\ours-A2/10 & Base & STEM & VEmbedding & \cellcolor{blue!8}\ours-A2/8 \\
    \midrule
    Mem & 0M & 755M & 94.4M & \cellcolor{blue!8}94.4M & \cellcolor{blue!8}188.7M & 0M & 1342.2M & 167.8M & \cellcolor{blue!8}167.8M & \cellcolor{blue!8}335.5M & 0M & 1409.3M & 469.8M & \cellcolor{blue!8}469.8M \\
    Wall & 1$\times$ & 1$\times$ & 1.03$\times$ & \cellcolor{blue!8}1.07$\times$ & \cellcolor{blue!8}1.07$\times$ & 1$\times$ & 1.09$\times$ & 1.04$\times$ & \cellcolor{blue!8}1.04$\times$ & \cellcolor{blue!8}1.08$\times$ & 1$\times$ & 1.07$\times$ & 1.06$\times$ & \cellcolor{blue!8}1.06$\times$ \\
    \midrule
    Train bpb & 0.8904 & 0.8683 & \underline{0.8671} & \cellcolor{blue!8}0.8673 & \cellcolor{blue!8}\textbf{0.8613} & 0.7807 & \textbf{0.7645} & 0.7710 & \cellcolor{blue!8}0.7704 & \cellcolor{blue!8}\underline{0.7675} & 0.7577 & 0.7484 & \textbf{0.7374} & \cellcolor{blue!8}\underline{0.7381} \\
    Val bpb & 0.8852 & 0.8633 & \underline{0.8620} & \cellcolor{blue!8}0.8623 & \cellcolor{blue!8}\textbf{0.8560} & 0.7750 & 0.7697 & 0.7651 & \cellcolor{blue!8}\underline{0.7647} & \cellcolor{blue!8}\textbf{0.7610} & 0.7594 & 0.7564 & \textbf{0.7427} & \cellcolor{blue!8}\underline{0.7461} \\
    CORE & 0.1382 & 0.1483 & 0.1530 & \cellcolor{blue!8}\underline{0.1587} & \cellcolor{blue!8}\textbf{0.1686} & 0.1988 & 0.2179 & \underline{0.2214} & \cellcolor{blue!8}0.2202 & \cellcolor{blue!8}\textbf{0.2286} & 0.2491 & \underline{0.2556} & 0.2530 & \cellcolor{blue!8}\textbf{0.2737} \\
    \midrule
    HSwag-0 & 34.11 & 35.81 & \underline{36.28} & \cellcolor{blue!8}36.15 & \cellcolor{blue!8}\textbf{36.59} & 48.00 & 47.65 & \underline{49.35} & \cellcolor{blue!8}49.11 & \cellcolor{blue!8}\textbf{49.79} & 50.12 & \underline{50.81} & \textbf{52.66} & \cellcolor{blue!8}\textbf{52.66} \\
    Jeop & 0.76 & \underline{2.36} & 0.90 & \cellcolor{blue!8}0.85 & \cellcolor{blue!8}\textbf{3.83} & 13.79 & 15.45 & \textbf{17.48} & \cellcolor{blue!8}16.34 & \cellcolor{blue!8}\underline{16.67} & 18.19 & 22.34 & \textbf{23.29} & \cellcolor{blue!8}\underline{22.82} \\
    ARC-E & 52.82 & 55.72 & \underline{56.82} & \cellcolor{blue!8}56.44 & \cellcolor{blue!8}\textbf{57.24} & 64.23 & 64.52 & \textbf{66.29} & \cellcolor{blue!8}65.03 & \cellcolor{blue!8}\underline{65.28} & 65.99 & \textbf{67.97} & 66.50 & \cellcolor{blue!8}\underline{67.09} \\
    ARC-C & 27.56 & \textbf{29.35} & \underline{28.50} & \cellcolor{blue!8}27.30 & \cellcolor{blue!8}27.65 & 34.04 & 35.41 & 36.26 & \cellcolor{blue!8}\underline{36.60} & \cellcolor{blue!8}\textbf{36.77} & 35.32 & 36.52 & \underline{38.05} & \cellcolor{blue!8}\textbf{38.31} \\
    COPA & 58.00 & \underline{65.00} & 64.00 & \cellcolor{blue!8}63.00 & \cellcolor{blue!8}\textbf{66.00} & 65.00 & 66.00 & 67.00 & \cellcolor{blue!8}\underline{68.00} & \cellcolor{blue!8}\textbf{70.00} & \textbf{72.00} & 69.00 & \underline{71.00} & \cellcolor{blue!8}68.00 \\
    CSQA & \textbf{31.61} & 27.19 & 22.11 & \cellcolor{blue!8}29.48 & \cellcolor{blue!8}\underline{30.71} & 19.74 & \textbf{22.44} & 20.72 & \cellcolor{blue!8}\underline{21.95} & \cellcolor{blue!8}20.31 & 22.36 & \underline{23.10} & 22.28 & \cellcolor{blue!8}\textbf{28.99} \\
    PIQA & 63.82 & 65.13 & \textbf{66.43} & \cellcolor{blue!8}\underline{66.00} & \cellcolor{blue!8}65.18 & 70.13 & 69.37 & 70.62 & \cellcolor{blue!8}\underline{70.73} & \cellcolor{blue!8}\textbf{71.60} & 70.78 & \textbf{71.98} & 71.00 & \cellcolor{blue!8}\underline{71.76} \\
    OBQA & 33.40 & \textbf{34.20} & 31.20 & \cellcolor{blue!8}\underline{34.00} & \cellcolor{blue!8}32.00 & \textbf{37.20} & 36.80 & \underline{37.00} & \cellcolor{blue!8}35.80 & \cellcolor{blue!8}36.00 & 37.60 & \underline{39.00} & \underline{39.00} & \cellcolor{blue!8}\textbf{39.60} \\
    LAMB & 27.58 & 28.68 & \underline{29.54} & \cellcolor{blue!8}29.38 & \cellcolor{blue!8}\textbf{30.72} & 39.41 & 38.19 & 39.94 & \cellcolor{blue!8}\underline{41.51} & \cellcolor{blue!8}\textbf{42.05} & \underline{43.16} & 42.03 & 42.98 & \cellcolor{blue!8}\textbf{43.45} \\
    HSwag & 33.27 & 35.47 & \textbf{36.23} & \cellcolor{blue!8}\underline{35.59} & \cellcolor{blue!8}\textbf{36.19} & 47.87 & 47.64 & \textbf{49.60} & \cellcolor{blue!8}\underline{49.20} & \cellcolor{blue!8}\textbf{49.50} & 50.58 & 51.01 & \underline{53.07} & \cellcolor{blue!8}\textbf{53.39} \\
    Winog & \underline{60.81} & 58.61 & 57.51 & \cellcolor{blue!8}\textbf{62.64} & \cellcolor{blue!8}60.44 & 65.20 & 62.64 & \underline{67.40} & \cellcolor{blue!8}66.67 & \cellcolor{blue!8}\textbf{68.50} & 67.03 & \textbf{71.43} & 69.23 & \cellcolor{blue!8}\underline{69.96} \\
    WinoG & \textbf{53.04} & \underline{52.88} & 52.09 & \cellcolor{blue!8}52.49 & \cellcolor{blue!8}51.93 & 54.14 & 54.54 & \underline{56.35} & \cellcolor{blue!8}56.04 & \cellcolor{blue!8}\textbf{56.75} & 55.96 & \underline{56.59} & \textbf{57.14} & \cellcolor{blue!8}56.04 \\
    LSAT & 23.91 & 23.91 & \underline{25.65} & \cellcolor{blue!8}\textbf{26.09} & \cellcolor{blue!8}\underline{25.65} & \textbf{27.83} & \underline{23.91} & 21.30 & \cellcolor{blue!8}23.04 & \cellcolor{blue!8}23.04 & 24.35 & \underline{24.78} & 24.35 & \cellcolor{blue!8}\textbf{26.09} \\
    SQuAD & 4.81 & 6.14 & \underline{13.61} & \cellcolor{blue!8}8.96 & \cellcolor{blue!8}\textbf{14.31} & 26.35 & 28.98 & \textbf{31.43} & \cellcolor{blue!8}\underline{30.57} & \cellcolor{blue!8}\underline{30.61} & 28.05 & 29.78 & \textbf{35.87} & \cellcolor{blue!8}\underline{35.62} \\
    CoQA & 10.12 & 12.31 & 13.52 & \cellcolor{blue!8}\underline{14.17} & \cellcolor{blue!8}\textbf{14.73} & 21.48 & 23.19 & \textbf{25.23} & \cellcolor{blue!8}\underline{24.14} & \cellcolor{blue!8}23.60 & 24.60 & 25.96 & \underline{28.18} & \cellcolor{blue!8}\textbf{28.50} \\
    BoolQ & \underline{55.14} & 54.01 & 53.67 & \cellcolor{blue!8}50.49 & \cellcolor{blue!8}\textbf{57.86} & \textbf{55.66} & 50.52 & 47.06 & \cellcolor{blue!8}44.65 & \cellcolor{blue!8}\underline{55.44} & \underline{59.88} & 56.39 & 47.55 & \cellcolor{blue!8}\textbf{63.39} \\
    BB-QAW & 31.48 & 37.23 & \underline{40.29} & \cellcolor{blue!8}\textbf{41.36} & \cellcolor{blue!8}39.13 & 53.54 & \underline{54.24} & 51.66 & \cellcolor{blue!8}53.64 & \cellcolor{blue!8}\textbf{54.54} & \underline{54.93} & \textbf{56.58} & 54.43 & \cellcolor{blue!8}53.89 \\
    BB-Dyck & 6.20 & 8.40 & \textbf{12.60} & \cellcolor{blue!8}\underline{10.60} & \cellcolor{blue!8}8.40 & 0.00 & 8.20 & \textbf{9.70} & \cellcolor{blue!8}\underline{8.40} & \cellcolor{blue!8}\underline{8.50} & \textbf{16.70} & \underline{14.10} & 13.20 & \cellcolor{blue!8}9.90 \\
    BB-CS & \textbf{43.11} & 38.33 & \textbf{43.03} & \cellcolor{blue!8}\underline{39.92} & \cellcolor{blue!8}39.77 & 0.00 & \textbf{43.03} & 34.24 & \cellcolor{blue!8}\underline{41.97} & \cellcolor{blue!8}18.26 & 40.98 & 36.89 & \textbf{45.45} & \cellcolor{blue!8}\underline{44.17} \\
    BB-Op & 11.90 & 10.95 & 12.38 & \cellcolor{blue!8}\textbf{16.19} & \cellcolor{blue!8}\underline{15.71} & 17.14 & 15.71 & \textbf{21.90} & \cellcolor{blue!8}18.57 & \cellcolor{blue!8}\underline{20.00} & 16.19 & 19.52 & \underline{22.38} & \cellcolor{blue!8}\textbf{23.81} \\
    BB-RCL & \underline{3.12} & 0.00 & 0.00 & \cellcolor{blue!8}\textbf{6.25} & \cellcolor{blue!8}\underline{3.12} & \underline{3.12} & \textbf{9.38} & 0.00 & \cellcolor{blue!8}0.00 & \cellcolor{blue!8}0.00 & \underline{0.00} & \textbf{3.12} & \underline{0.00} & \cellcolor{blue!8}\underline{0.00} \\
    BB-LI & 24.96 & \textbf{25.88} & 24.97 & \cellcolor{blue!8}25.52 & \cellcolor{blue!8}\underline{25.72} & \textbf{26.30} & 25.07 & \underline{25.70} & \cellcolor{blue!8}\underline{25.64} & \cellcolor{blue!8}\underline{25.62} & \underline{25.50} & \textbf{25.87} & \underline{25.48} & \cellcolor{blue!8}25.35 \\
    \bottomrule
  \end{tabular}%
  }
\end{table*}

\FloatBarrier

\subsubsection{\texorpdfstring{Token-Indexed Memory Baselines}{Token-Indexed Memory Baselines}}
\label{sec:appendix-token-memory-baselines}

Table~\ref{tab:appendix-token-memory-adaptations} compares \ours{} with token-indexed memory baselines, including adaptations of JTok and JTok-M~\citep{yang2026jtok} and MoVE-Input~\citep{uppal2026move} to our backbone and training setup. Bigram gives the lowest validation bpb, while \ours{} gives the highest CORE-22.

\begin{table*}[!htbp]
  \caption{Token-indexed memory comparisons on d24/w1536/h12, ClimbMix, seed~42, at approximately $6\times10^{19}$ training FLOPs. Bigram is the 2-gram baseline used in the main comparisons. Canonical Engram, JTok, JTok-M, and MoVE-Input are adapted to our backbone at approximately matched parameter counts; their architectures and training settings differ from the originals. Appendix~\ref{sec:appendix-baseline-implementation} describes the Engram adaptation.}
  \label{tab:appendix-token-memory-adaptations}
  \centering
  \small
  \setlength{\tabcolsep}{12pt}
  \renewcommand{\arraystretch}{1.12}
  \begin{tabular}{@{}lrrr@{}}
    \toprule
    Method & Parameters (M) & Validation bpb $\downarrow$ & CORE-22 $\uparrow$ \\
    \midrule
    Bigram & 1,384.1 & \textbf{0.6943} & 0.2871 \\
    Canonical Engram adaptation & 1,389.6 & 0.7029 & 0.2698 \\
    JTok adaptation & 1,384.1 & 0.6974 & 0.2764 \\
    JTok-M adaptation & 1,384.1 & 0.7003 & 0.2756 \\
    MoVE-Input adaptation & 1,386.7 & 0.7016 & 0.2744 \\
    \rowcolor{blue!8}
    \ours{} & 1,386.8 & 0.6990 & \textbf{0.2941} \\
    \bottomrule
  \end{tabular}
\end{table*}

\FloatBarrier

\subsection{\texorpdfstring{Scaling and Ablations}{Scaling and Ablations}}
\label{sec:appendix-scaling-ablations}

\subsubsection{\texorpdfstring{Memory-size scaling}{Memory-size scaling}}
\label{sec:appendix-memory-scaling}

\begin{figure}[!ht]
  \centering
  \includegraphics[width=\linewidth]{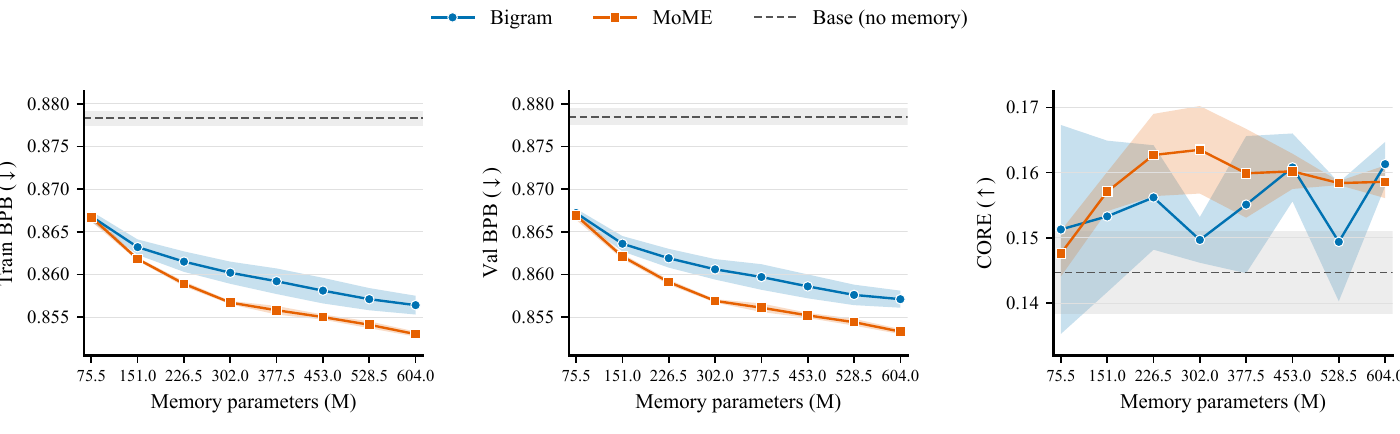}
  \caption{Full memory-size scaling sweep on the 12-layer nanochat backbone (3e18 FLOPs, 3-seed mean $\pm$ std). From left to right: training bpb, validation bpb, and CORE vs.\ memory parameter count for Bigram and \ours{} for block counts $\{3,6,9,12,15,18,21,24\}$. The dashed line and shaded band are the no-memory Base model from Table~\ref{tab:pretrain-12layer}.}
  \label{fig:scaling-curve-appendix}
\end{figure}

Table~\ref{tab:appendix-memory-scaling} and Figure~\ref{fig:scaling-curve-appendix} compare Bigram and \ours{} as memory size increases on the nanochat d12 backbone, with the no-memory Base model from Table~\ref{tab:pretrain-12layer} as a reference.

% Memory-size scaling sweep: Engram vs. MoME at block counts {3,6,9,12,15,18,21,24}
% memory blocks, 12-layer nanochat backbone, 3e18 training FLOPs, 3-seed mean
% +- std. Base-model row imported from Table~\ref{tab:pretrain-12layer}.
\begin{table}[!ht]
  \caption{Memory-size scaling sweep on the 12-layer nanochat backbone trained at 3e18 FLOPs ($\sim$3.3B tokens), with block counts $\{3, 6, 9, 12, 15, 18, 21, 24\}$. \textit{Blocks} is the memory expansion factor; each block adds $\approx 25.2$M parameters to the memory table. Values are 3-seed means $\pm$ std. The Base-model row is the no-memory baseline imported from Table~\ref{tab:pretrain-12layer}. Bigram and \ours{} share the same memory budget at each block count, so the rows are directly comparable. Figure~\ref{fig:scaling-curve} plots the val bpb column vs.\ memory.}
  \label{tab:appendix-memory-scaling}
  \centering
  \scriptsize
  \setlength{\tabcolsep}{5pt}
  \renewcommand{\arraystretch}{1.20}
  \begin{tabular}{@{}l r r r r r r@{}}
    \toprule
    Method & Blocks & Mem (M) & Train BPB $\downarrow$ & Val BPB $\downarrow$ & CORE $\uparrow$ & Wall (min) \\
    \midrule
    Base   & --   & 0     & \metricstd{0.8783}{0.0009} & \metricstd{0.8785}{0.0010} & \metricstd{0.1447}{0.0064} & --                       \\
    \midrule
    Bigram & 3    & 75.5  & \metricstd{0.8668}{0.0006} & \metricstd{0.8672}{0.0005} & \metricstd{0.1513}{0.0160} & \metricstd{106.72}{0.70} \\
    Bigram & 6    & 151.0 & \metricstd{0.8632}{0.0009} & \metricstd{0.8636}{0.0009} & \metricstd{0.1533}{0.0116} & \metricstd{108.80}{0.47} \\
    Bigram & 9    & 226.5 & \metricstd{0.8615}{0.0012} & \metricstd{0.8619}{0.0011} & \metricstd{0.1562}{0.0080} & \metricstd{110.55}{0.50} \\
    Bigram & 12   & 302.0 & \metricstd{0.8602}{0.0013} & \metricstd{0.8606}{0.0012} & \metricstd{0.1497}{0.0035} & \metricstd{112.49}{0.38} \\
    Bigram & 15   & 377.5 & \metricstd{0.8592}{0.0015} & \metricstd{0.8597}{0.0015} & \metricstd{0.1551}{0.0105} & \metricstd{114.35}{0.43} \\
    Bigram & 18   & 453.0 & \metricstd{0.8581}{0.0015} & \metricstd{0.8586}{0.0014} & \metricstd{0.1608}{0.0052} & \metricstd{116.34}{0.60} \\
    Bigram & 21   & 528.5 & \metricstd{0.8571}{0.0013} & \metricstd{0.8576}{0.0012} & \metricstd{0.1494}{0.0091} & \metricstd{118.28}{0.40} \\
    Bigram & 24   & 604.0 & \metricstd{0.8564}{0.0011} & \metricstd{0.8571}{0.0010} & \metricstd{0.1613}{0.0034} & \metricstd{120.17}{0.45} \\
    \midrule
    \ours{} & 3    & 75.5  & \metricstd{0.8667}{0.0005} & \metricstd{0.8669}{0.0005} & \metricstd{0.1476}{0.0035} & \metricstd{109.77}{0.82} \\
    \ours{} & 6    & 151.0 & \metricstd{0.8618}{0.0002} & \metricstd{0.8621}{0.0003} & \metricstd{0.1571}{0.0030} & \metricstd{111.89}{0.59} \\
    \ours{} & 9    & 226.5 & \metricstd{0.8589}{0.0003} & \metricstd{0.8591}{0.0003} & \metricstd{0.1627}{0.0063} & \metricstd{113.48}{0.59} \\
    \ours{} & 12   & 302.0 & \metricstd{0.8567}{0.0002} & \metricstd{0.8569}{0.0002} & \metricstd{0.1635}{0.0067} & \metricstd{115.68}{0.54} \\
    \ours{} & 15   & 377.5 & \metricstd{0.8558}{0.0005} & \metricstd{0.8561}{0.0005} & \metricstd{0.1599}{0.0068} & \metricstd{117.59}{0.57} \\
    \ours{} & 18   & 453.0 & \metricstd{0.8550}{0.0003} & \metricstd{0.8552}{0.0003} & \metricstd{0.1602}{0.0027} & \metricstd{119.76}{0.51} \\
    \ours{} & 21   & 528.5 & \metricstd{0.8541}{0.0004} & \metricstd{0.8544}{0.0004} & \metricstd{0.1584}{0.0003} & \metricstd{121.85}{0.60} \\
    \ours{} & 24   & 604.0 & \metricstd{0.8530}{0.0003} & \metricstd{0.8533}{0.0003} & \metricstd{0.1586}{0.0025} & \metricstd{123.70}{0.51} \\
    \bottomrule
  \end{tabular}
\end{table}

\FloatBarrier

\FloatBarrier

\subsubsection{\texorpdfstring{Context-Conditioned Routing}{Context-Conditioned Routing}}
\label{sec:appendix-context-routing-controls}

Table~\ref{tab:appendix-context-routing-controls} compares routing rules while retaining multiple trainable slots at nearly matched parameter and compute budgets. Hidden-state routing improves validation bpb and CORE-22 over learned token-only and fixed-random routing.

\begin{table*}[!htbp]
  \caption{Routing controls on d24/w1536/h12, ClimbMix, seed~42, at approximately $6\times10^{19}$ training FLOPs. All variants retain multiple slots at nearly matched parameter budgets; only hidden-state routing uses context. Both delta columns report the control value minus the \ours{} value.}
  \label{tab:appendix-context-routing-controls}
  \centering
  \small
  \setlength{\tabcolsep}{8pt}
  \renewcommand{\arraystretch}{1.12}
  \resizebox{\textwidth}{!}{%
  \begin{tabular}{@{}lccrrrr@{}}
    \toprule
    Variant & Context-conditioned & Parameters (M) & Validation bpb $\downarrow$ & $\Delta$ bpb & CORE-22 $\uparrow$ & $\Delta$ CORE \\
    \midrule
    \rowcolor{blue!8}
    \ours{}, hidden-state routing & Yes & 1,386.8 & \textbf{0.6990} & --- & \textbf{0.2941} & --- \\
    \ours{}, learned token-only routing & No & 1,388.8 & 0.7044 & +0.0054 & 0.2853 & -0.0088 \\
    \ours{}, fixed-random routing & No & 1,384.1 & 0.7018 & +0.0028 & 0.2768 & -0.0173 \\
    \bottomrule
  \end{tabular}%
  }
\end{table*}

\FloatBarrier

\subsubsection{\texorpdfstring{Dense Scaling Controls}{Dense Scaling Controls}}
\label{sec:appendix-dense-scaling-controls}

Table~\ref{tab:appendix-dense-scaling-controls} compares \ours{} with pure dense backbones at the original width, at a width selected to match measured inference latency, and at approximately matched total parameters. Under the same training-FLOP budget, \ours{} has the lowest bpb on the native validation set and all three shifted-domain evaluations and the highest CORE-22.

\begin{table*}[!htbp]
  \caption{Dense-scaling controls on ClimbMix, seed~42, at approximately $6\times10^{19}$ training FLOPs. The latency-matched width is selected using measured inference latency before training; the matched-total dense model is compute-matched rather than token-matched because its larger dense backbone processes fewer tokens under the fixed FLOP budget. Validation bpb uses the native evaluator, while the three additional corpora use the standardized document-reset protocol.}
  \label{tab:appendix-dense-scaling-controls}
  \centering
  \scriptsize
  \setlength{\tabcolsep}{5pt}
  \renewcommand{\arraystretch}{1.12}
  \resizebox{\textwidth}{!}{%
  \begin{tabular}{@{}lrrrrrr@{}}
    \toprule
    Model & Parameters (M) & Validation bpb $\downarrow$ & FineWeb-Edu bpb $\downarrow$ & enwik9 bpb $\downarrow$ & Shakespeare bpb $\downarrow$ & CORE-22 $\uparrow$ \\
    \midrule
    Pure dense d24/w1536 & 780 & 0.7042 & 0.8105 & 0.9775 & 1.5610 & 0.2843 \\
    Latency-matched dense d24/w1728 & 973 & 0.6998 & 0.8097 & 0.9757 & 1.5632 & 0.2773 \\
    Matched-total dense d25/w2048 & 1,393 & 0.7033 & 0.8092 & 0.9780 & 1.5584 & 0.2871 \\
    \rowcolor{blue!8}
    \ours{} d24/w1536 & 1,387 & \textbf{0.6990} & \textbf{0.8045} & \textbf{0.9712} & \textbf{1.5531} & \textbf{0.2941} \\
    \bottomrule
  \end{tabular}%
  }
\end{table*}

\FloatBarrier

\subsubsection{\texorpdfstring{Test-time memory use across CORE domains}{Test-time memory use across CORE domains}}
\label{sec:appendix-memory-domain-audit}

We evaluate the step-100{,}000 $d24$ checkpoint used in Figure~\ref{fig:main-model-analysis}. The model has 12 odd-numbered memory layers, 12 routing heads per memory layer, 12 slots per head, and top-2 routing. The ClimbMix reference contains 1{,}048{,}576 executed token positions; the CORE audit evaluates all examples in the 22-task bundle. For multiple-choice tasks, repeated candidate executions are counted because they are part of the model's actual test-time computation. Cross-task summaries give every task equal weight.

\begin{figure}[!ht]
  \centering
  \includegraphics[width=\linewidth]{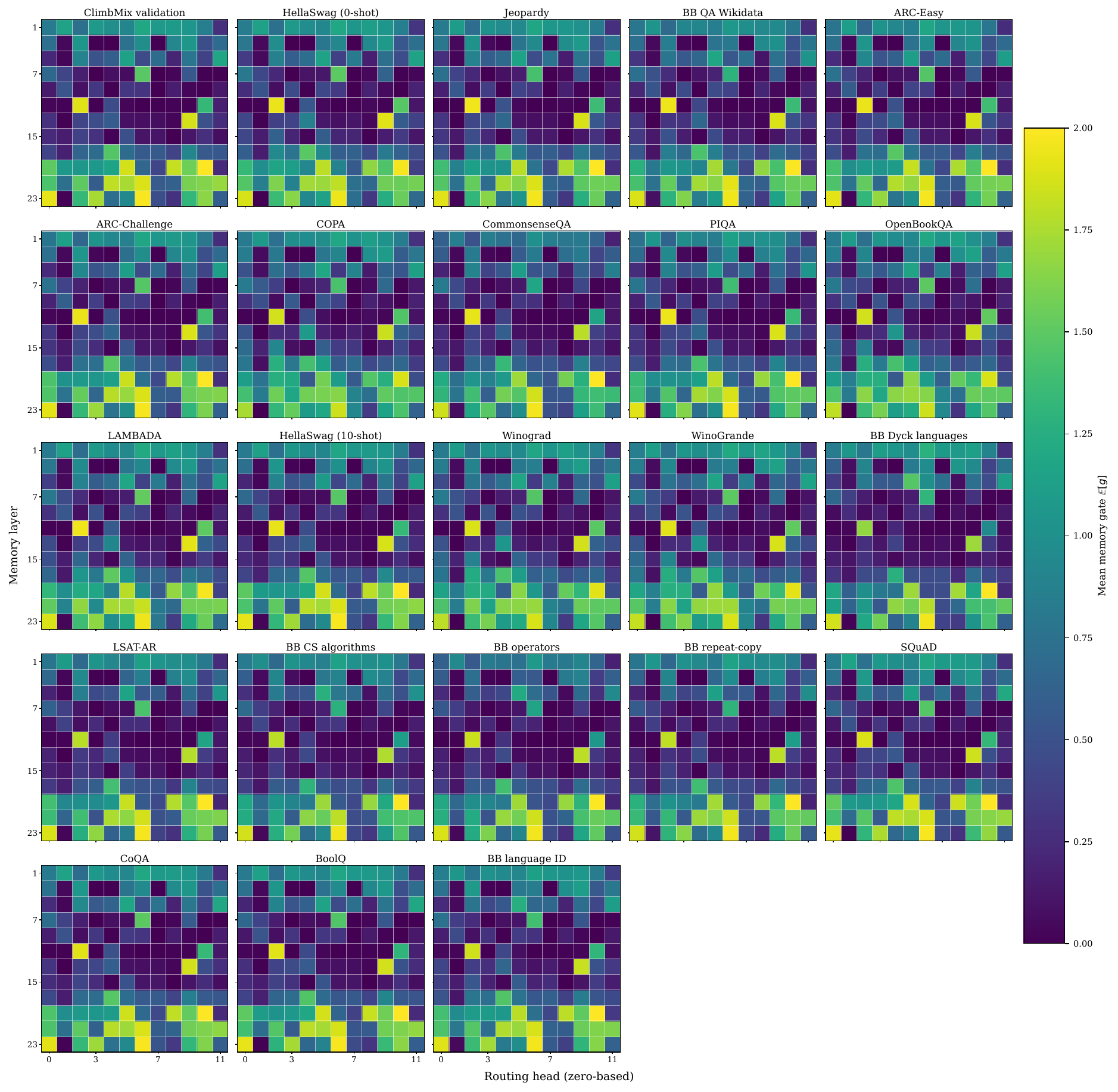}
  \caption{Mean injection gates for ClimbMix validation and all 22 CORE tasks. Each panel shows 12 memory layers by 12 routing heads on the same $[0,2]$ color scale.}
  \label{fig:appendix-memory-gate-by-domain}
\end{figure}
\FloatBarrier

Figure~\ref{fig:appendix-memory-gate-by-domain} shows similar gate patterns across CORE tasks, with substantial variation between heads within each layer. Some heads have nearly closed injection gates, while others have gates near their upper bound; layer averages obscure these differences in injection strength.

\begin{figure}[!ht]
  \centering
  \includegraphics[width=\linewidth]{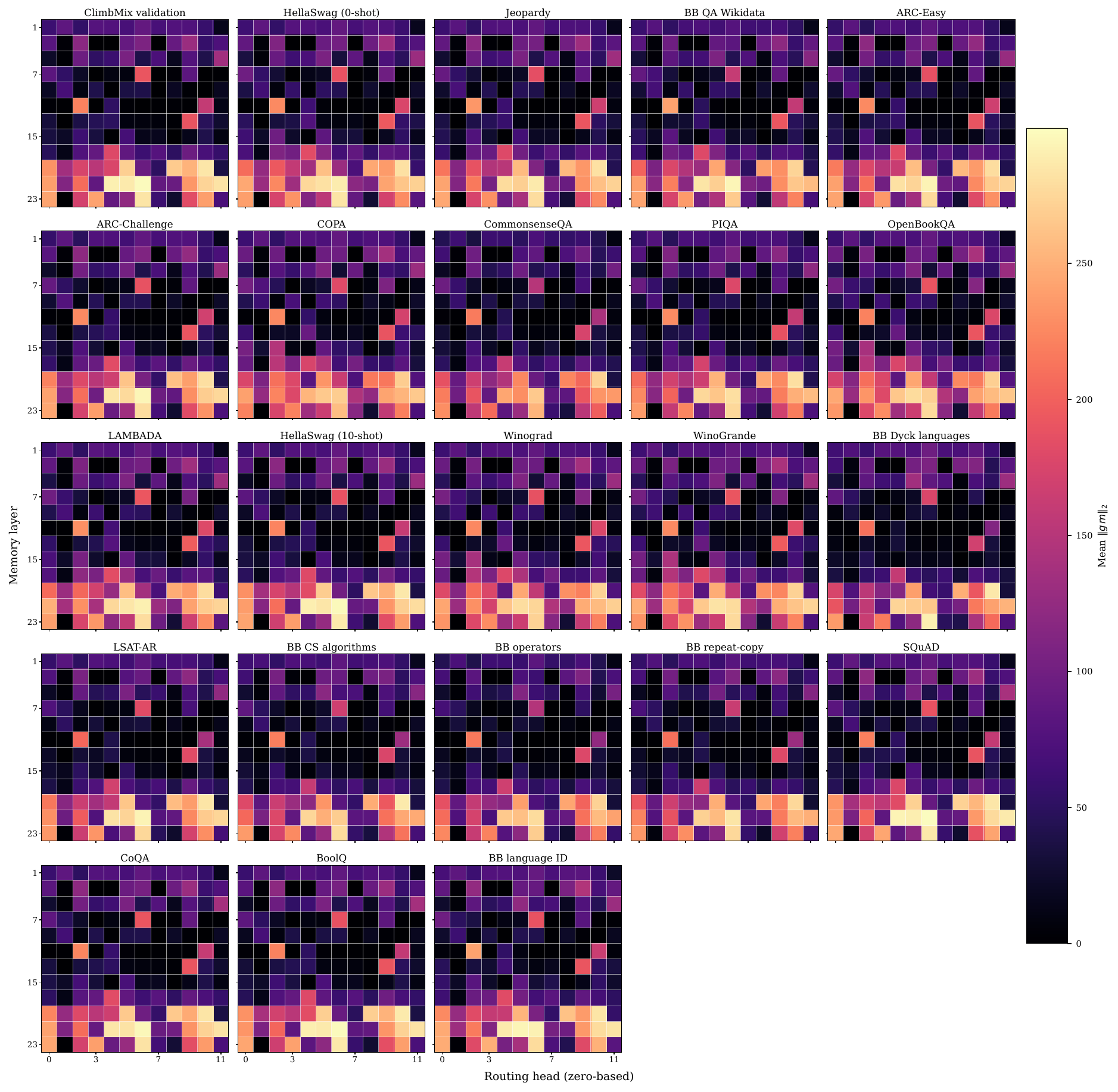}
  \caption{Actual injected memory-branch magnitude for ClimbMix validation and all 22 CORE tasks. Each cell reports $\mathbb{E}[\|\gamma\mathbf{m}\|_2]$; all panels share one absolute color scale.}
  \label{fig:appendix-memory-effective-l2-by-domain}
\end{figure}
\FloatBarrier

The injected-memory norms in Figure~\ref{fig:appendix-memory-effective-l2-by-domain} are largest in the late layers on both ClimbMix and the CORE tasks. These norms measure injection magnitude; they do not establish how much the memory contributes to predictions.

\subsubsection{\texorpdfstring{Router slot utilization across domains}{Router slot utilization across domains}}
\label{sec:appendix-slot-utilization}

We evaluate the top-2 router of the 100B-token checkpoint on 1{,}048{,}576 source-token positions from each of ClimbMix validation, FineWeb-Edu validation, enwik9, and Shakespeare. Beginning-of-sequence and padding positions are excluded. Statistics are computed at each of the 12 memory layers and 12 routing heads, then averaged over the 144 sites.

For a slot distribution $p$, the effective slot count is $M_{\mathrm{eff}}=\exp(H(p))$. We compute it using either unweighted top-2 membership or routing weights. We also compute the membership distribution separately for each token ID observed at least 32 times, to assess whether repeated occurrences use different slot pairs.

\begin{table*}[!htbp]
  \caption{Router slot-utilization audit for the 100B-token \ours{} checkpoint. Effective counts are macro-averaged over 12 memory layers and 12 routing heads. Per-token counts retain token IDs with at least 32 occurrences. Inactive cells count layer--head--slot combinations never selected on the evaluated corpus. JSD compares each site's routing-weight distribution with ClimbMix. These statistics measure router concentration, not learned memory-table coverage.}
  \label{tab:appendix-slot-utilization}
  \centering
  \scriptsize
  \setlength{\tabcolsep}{5pt}
  \renewcommand{\arraystretch}{1.12}
  \resizebox{\textwidth}{!}{%
  \begin{tabular}{@{}lrrrrrr@{}}
    \toprule
    Domain & Marginal effective slots & Weight-effective slots & Largest-slot mass & Per-token effective slots & Inactive site--slot cells & Mean/max JSD \\
    \midrule
    ClimbMix validation & 4.614 & 3.265 & 61.66\% & 3.550 & 4 / 1,728 & 0 / 0 \\
    FineWeb-Edu validation & 4.576 & 3.222 & 61.85\% & 3.491 & 9 / 1,728 & 0.0026 / 0.0282 \\
    enwik9 & 4.528 & 3.209 & 60.95\% & 3.353 & 12 / 1,728 & 0.0235 / 0.1412 \\
    Shakespeare & 4.230 & 2.959 & 64.01\% & 3.082 & 42 / 1,728 & 0.0305 / 0.2975 \\
    \bottomrule
  \end{tabular}%
  }
\end{table*}

\FloatBarrier

Table~\ref{tab:appendix-slot-utilization} shows that routing mass concentrates on a subset of slots, while repeated occurrences of the same token use different slot pairs across contexts. In every domain, the token-conditional effective slot count exceeds the number of slots selected per occurrence. Routing distributions differ more from ClimbMix on enwik9 and Shakespeare than on FineWeb-Edu.

\subsection{\texorpdfstring{Full Inference Latency Results}{Full Inference Latency Results}}
\label{sec:appendix-inference-latency}

Table~\ref{tab:appendix-inference-latency} reports the full latency sweep
corresponding to the compact main-text summary in
Table~\ref{tab:inference-latency}.

\begin{table*}[!htbp]
  \caption{Full inference latency overhead of memory-augmented value retrieval across backbone scales. Baseline is the absolute latency in milliseconds; memory columns report absolute deltas relative to the baseline, with percentage overhead in parentheses. $L_{\mathrm{mem}}$ denotes the number of memory-injected layers in this benchmark. \emph{\ours{} (hidden-state injection)} denotes the \ours{} variant that gates from the input and injects back into the input stream. It is an injection-site ablation of \ours{}, not the canonical Engram model.}
  \label{tab:appendix-inference-latency}
  \centering
  \scriptsize
  \setlength{\tabcolsep}{4pt}
  \renewcommand{\arraystretch}{1.12}
  \resizebox{\textwidth}{!}{%
  \begin{tabular}{@{}l r r r r r r@{}}
    \toprule
    Arch & $L_{\mathrm{mem}}$ & Baseline ms & VE & Bigram & \ours{} & \ours{} (hidden-state injection) \\
    \midrule
    \texttt{mobilellm350} & 2 & 3.069 & $+0.017$ ($+0.55\%$) & $-0.036$ ($-1.18\%$) & $+0.240$ ($+7.82\%$) & $+0.247$ ($+8.05\%$) \\
    \texttt{mobilellm350} & 3 & 4.526 & $+0.025$ ($+0.55\%$) & $+0.073$ ($+1.61\%$) & $+0.249$ ($+5.50\%$) & $+0.299$ ($+6.61\%$) \\
    \texttt{mobilellm1b} & 2 & 5.065 & $+0.024$ ($+0.47\%$) & $-0.016$ ($-0.31\%$) & $+0.277$ ($+5.47\%$) & $+0.272$ ($+5.37\%$) \\
    \texttt{mobilellm1b} & 3 & 7.481 & $+0.035$ ($+0.47\%$) & $+0.151$ ($+2.02\%$) & $+0.295$ ($+3.95\%$) & $+0.317$ ($+4.24\%$) \\
    \texttt{qwen3\_0p6b} & 2 & 4.768 & $+0.053$ ($+1.11\%$) & $-0.038$ ($-0.80\%$) & $+0.377$ ($+7.90\%$) & $+0.307$ ($+6.44\%$) \\
    \texttt{qwen3\_0p6b} & 3 & 7.065 & $+0.050$ ($+0.71\%$) & $+0.089$ ($+1.26\%$) & $+0.378$ ($+5.35\%$) & $+0.361$ ($+5.11\%$) \\
    \texttt{qwen3\_4b} & 2 & 21.665 & $+0.099$ ($+0.46\%$) & $+0.088$ ($+0.41\%$) & $+0.509$ ($+2.35\%$) & $+0.996$ ($+4.60\%$) \\
    \texttt{qwen3\_4b} & 3 & 32.532 & $+0.129$ ($+0.40\%$) & $+0.236$ ($+0.73\%$) & $+0.503$ ($+1.55\%$) & $+1.125$ ($+3.46\%$) \\
    \texttt{qwen3\_8b} & 2 & 37.906 & $+0.079$ ($+0.21\%$) & $+0.025$ ($+0.07\%$) & $+0.590$ ($+1.56\%$) & $+0.846$ ($+2.23\%$) \\
    \texttt{qwen3\_8b} & 3 & 56.724 & $+0.085$ ($+0.15\%$) & $+0.267$ ($+0.47\%$) & $+0.592$ ($+1.04\%$) & $+0.948$ ($+1.67\%$) \\
    \bottomrule
  \end{tabular}%
  }
\end{table*}

\FloatBarrier

\clearpage
\begingroup
\raggedbottom
\subsection{\texorpdfstring{Supplementary Semantic-Routing Examples}{Supplementary Semantic-Routing Examples}}
\label{sec:appendix-semantic-routes}

We show selected examples for \emph{apple}, \emph{bank}, \emph{cell}, and \emph{python} on nanochat d24 and Qwen 0.6B. Each panel compares two prompts with the same target-word sense and a third prompt with a different sense.

\input{tables/appendix_semantic_routes}
\clearpage
\endgroup

\subsection{\texorpdfstring{Quantifying Sense-Sensitive Routing on WiC}{Quantifying Sense-Sensitive Routing on WiC}}
\label{sec:appendix-wic-routing}

\paragraph{Dataset and filtering.}
We complement the curated examples above with a quantitative probe on the union of the train and test splits of WiC~\citep{pilehvar2019wic}; no WiC examples are used to fit or tune the model or a probe. Each WiC item contains two natural sentences with the same marked target word and a binary label: $T$ if the target has the same sense in both sentences and $F$ otherwise. Starting from all 6,828 pairs (5,428 train and 1,400 test), we retain a \emph{strict-natural} pair only when the marked surface in both sentences matches the canonical target case-insensitively, each occurrence is represented by exactly one tokenizer token, and the two occurrences have the same token ID. This leaves 3,206 pairs (2,655 train and 551 test).

Because the memory router is causal and can only use the prefix preceding the target, we further require both marked occurrences to lie strictly after the midpoint of their tokenized sentence. Let $p(x)$ denote the target position in a sequence that includes the beginning-of-sequence token and let $L(x)$ denote the corresponding sequence length. We retain occurrence $x$ when
\begin{equation}
  r(x)=\frac{p(x)-\tfrac{1}{2}}{L(x)-1}>\frac{1}{2},
  \label{eq:wic-context-position}
\end{equation}
and retain a pair only if both occurrences pass. The resulting evaluation set contains 670 pairs (549 train and 121 test; 354 $T$ and 316 $F$), 362 unique target-token IDs, and 1,340 target occurrences. The minimum preceding-context length has a median of 5 tokens and an interquartile range of 4--6; 547 pairs have at least four preceding tokens. The retained set is heavily noun-skewed (634 noun pairs and 36 verb pairs).

\paragraph{Per-head metrics.}
We analyze routing in the $d24$ \ours{} model at each layer--head site $s=(\ell,h)$. The model has memory at the 12 odd-numbered layers $\{1,3,\ldots,23\}$, with 12 routing heads per layer and top-2 selection from 12 slots. For WiC pair $i$, let $\mathbf{q}_{i1,s}$ and $\mathbf{q}_{i2,s}$ be the full routing distributions for its two target occurrences, and let $\mathbf{m}_{i,s}=(\mathbf{q}_{i1,s}+\mathbf{q}_{i2,s})/2$. We compute the base-2 Jensen--Shannon divergence
\begin{equation}
  d_{i,s}
  =
  \frac{1}{2}\mathrm{KL}_{2}\!\left(\mathbf{q}_{i1,s}\middle\|\mathbf{m}_{i,s}\right)
  +
  \frac{1}{2}\mathrm{KL}_{2}\!\left(\mathbf{q}_{i2,s}\middle\|\mathbf{m}_{i,s}\right),
  \qquad
  \Delta\mathrm{JSD}_{s}
  =
  \mathbb{E}[d_{i,s}\mid F]-\mathbb{E}[d_{i,s}\mid T].
  \label{eq:wic-head-jsd}
\end{equation}
Positive $\Delta\mathrm{JSD}_{s}$ indicates greater average routing divergence for different-sense pairs than for same-sense pairs.

We also measure chance-corrected decision overlap following \citet{olson-etal-2025-probing}. Let $A_{i1,s}$ and $A_{i2,s}$ be the executed top-$K$ slot sets, let $o_{i,s}=|A_{i1,s}\cap A_{i2,s}|$, and let $M$ be the number of slots. We compute
\begin{equation}
  \widetilde O_{i,s}
  =
  \frac{o_{i,s}-K^2/M}{K-K^2/M},
  \qquad
  \Delta\widetilde O_s
  =
  \mathbb{E}[\widetilde O_{i,s}\mid T]
  -
  \mathbb{E}[\widetilde O_{i,s}\mid F].
  \label{eq:wic-head-overlap}
\end{equation}
Positive $\Delta\widetilde O_s$ indicates greater slot overlap for same-sense pairs, after correcting for uniform-random top-$K$ selection.

\begin{figure}[t]
  \centering
  \includegraphics[width=\linewidth]{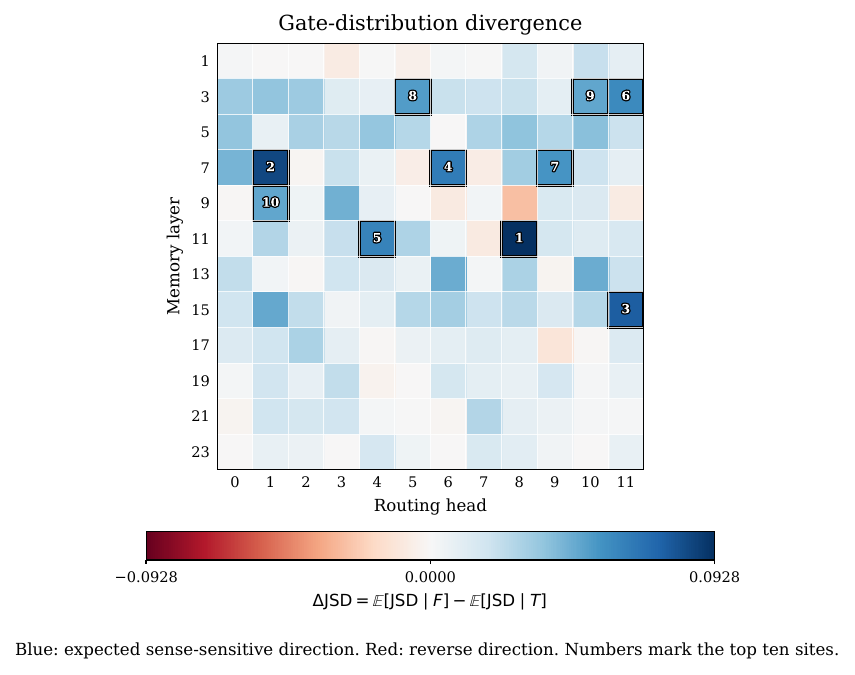}
  \caption{Routing-distribution differences on the 670 filtered WiC pairs. Cells correspond to layer--head sites. Blue indicates greater divergence for different-sense pairs ($\Delta\mathrm{JSD}_{s}>0$), and red indicates the reverse. Outlined numbers rank the ten largest positive effects. Head indices are zero-based.}
  \label{fig:appendix-wic-per-head-routing}
\end{figure}
\FloatBarrier

\paragraph{Results.}
In Figure~\ref{fig:appendix-wic-per-head-routing}, different-sense pairs have greater routing divergence at 117 of 144 sites. Most of the largest differences occur in the middle layers.

Layer~11/head~8 has the largest effect, with $\Delta\mathrm{JSD}=0.0928$ and a pair-stratified bootstrap 95\% interval of $[0.0489,0.1357]$. All ten largest effects have pointwise bootstrap intervals above zero.

Decision overlap is greater for same-sense pairs at most sites. The largest effect again occurs at layer~11/head~8, with $\Delta\widetilde O=0.0939$ and a pair-stratified bootstrap 95\% interval of $[0.0360,0.1520]$.

\begin{figure}[t]
  \centering
  \includegraphics[width=\linewidth]{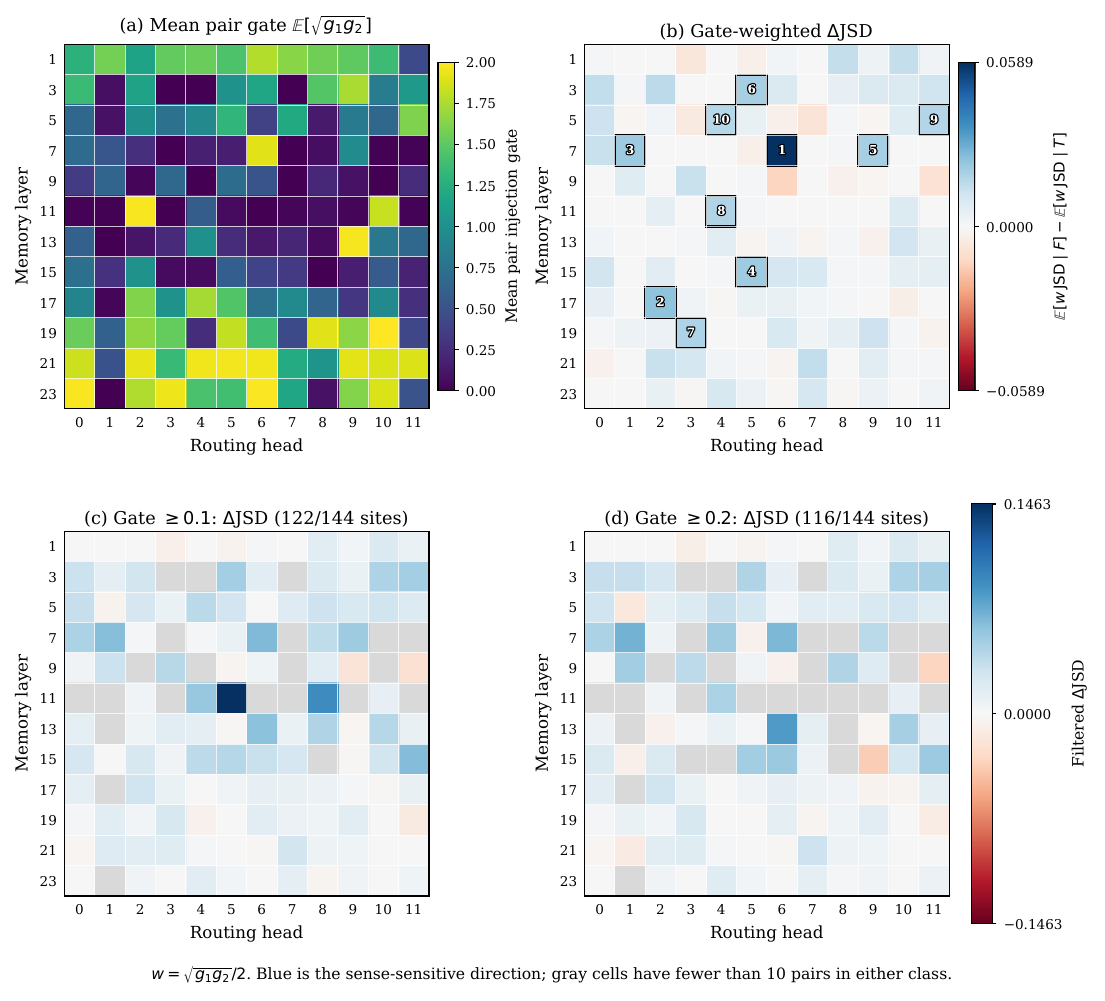}
  \caption{Injection-gate sensitivity analysis on the same WiC pairs as Figure~\ref{fig:appendix-wic-per-head-routing}. (a) Mean unnormalized pair gate $\mathbb{E}[\sqrt{\gamma_{i1,s}\gamma_{i2,s}}]$. (b) Gate-weighted effect from Equation~\ref{eq:wic-gate-weighted-jsd}, with the ten largest positive effects ranked. (c--d) Unweighted $\Delta\mathrm{JSD}_{s}$ for pairs whose two injection gates exceed $0.1$ or $0.2$, using a shared color scale. Gray cells have fewer than ten retained pairs in either class. Blue indicates positive effects; head indices are zero-based.}
  \label{fig:appendix-wic-gate-aware-routing}
\end{figure}

\paragraph{Injection-gate-aware sensitivity.}
To account for injection strength, we measure the per-head injection gate $\gamma_{ij,s}\in[0,2]$ for occurrence $j\in\{1,2\}$ of pair $i$ at site $s$. We weight each pair by $w_{i,s}=\sqrt{\gamma_{i1,s}\gamma_{i2,s}}/2\in[0,1]$, which decreases when either occurrence has a small injection gate, and compute
\begin{equation}
  \Delta\mathrm{JSD}^{\mathrm{gate}}_{s}
  =
  \mathbb{E}[w_{i,s}d_{i,s}\mid F]
  -
  \mathbb{E}[w_{i,s}d_{i,s}\mid T].
  \label{eq:wic-gate-weighted-jsd}
\end{equation}
We apply the weights before computing the class means. We also recompute both routing metrics after retaining only pairs with $\min(\gamma_{i1,s},\gamma_{i2,s})\geq\tau$, for $\tau\in\{0.1,0.2,0.25\}$. Thresholded cells with fewer than ten pairs in either sense class are shown in gray. Figure~\ref{fig:appendix-wic-gate-aware-routing} shows the first two JSD thresholds; Figure~\ref{fig:main-model-analysis} reports the unfiltered effects for both metrics.

\FloatBarrier

Accounting for injection strength changes which heads have the largest effects (Figure~\ref{fig:appendix-wic-gate-aware-routing}). Layer~11/head~8 has the largest unweighted routing difference but a nearly closed injection gate. Layer~7/head~6, which has a nearly open gate, has the largest gate-weighted effect.

At layer~7/head~6, $\Delta\mathrm{JSD}^{\mathrm{gate}}=0.0589$, with a pair-stratified bootstrap 95\% interval of $[0.0167,0.1016]$. Fifteen sites have pointwise weighted intervals above zero. The unweighted metric measures routing divergence; the weighted metric shows how that divergence changes when pairs with small injection gates contribute less.

At the strictest gate threshold, $\tau=0.25$, most retained sites still show greater routing divergence for different-sense pairs and greater slot overlap for same-sense pairs.

\paragraph{Scope and limitations.}
Routing differences are associated with word sense, including at heads with open injection gates, but this analysis does not measure their effect on predictions. Only 81 target-token groups contain both sense labels, so the pooled comparison may reflect differences in lexical composition despite holding the target token fixed within each pair. The analysis combines WiC train and test examples and does not estimate held-out generalization. Bootstrap intervals are pointwise, and the largest effects are selected on the same examples without correction across the 144 sites. Confirmatory testing would require separate data with repeated target types and a site-wise label-permutation test with max-$T$ or false-discovery-rate correction.

% \newpage
% \input{secs/lessons_learned}

\end{document}